\documentclass[letterpaper]{article} % DO NOT CHANGE THIS
\usepackage{aaai2026}  % DO NOT CHANGE THIS
\usepackage{times}  % DO NOT CHANGE THIS
\usepackage{helvet}  % DO NOT CHANGE THIS
\usepackage{courier}  % DO NOT CHANGE THIS
\usepackage[hyphens]{url}  % DO NOT CHANGE THIS
\usepackage{graphicx} % DO NOT CHANGE THIS
\usepackage{natbib}  % DO NOT CHANGE THIS AND DO NOT ADD ANY OPTIONS TO IT
\usepackage{caption} % DO NOT CHANGE THIS AND DO NOT ADD ANY OPTIONS TO IT
\usepackage{algorithm}
\usepackage{algorithmic}

\usepackage{newfloat}
\usepackage{listings}
\DeclareCaptionStyle{ruled}{labelfont=normalfont,labelsep=colon,strut=off} % DO NOT CHANGE THIS
\floatstyle{ruled}
\newfloat{listing}{tb}{lst}{}
\floatname{listing}{Listing}
\title{HLSMAC: A New StarCraft Multi-Agent Challenge for High-Level Strategic Decision-Making}
\author{
    Xingxing Hong\textsuperscript{\rm 1},
    Yungong Wang\textsuperscript{\rm 2},
    Dexin Jin\textsuperscript{\rm 3},
    Ye Yuan\textsuperscript{\rm 4},
    Ximing Huang\textsuperscript{\rm 1},
    Zijian Wu\textsuperscript{\rm 1},
    Wenxin Li\textsuperscript{\rm 1}
}
\affiliations {
    \textsuperscript{\rm 1}Peking University\\
    \textsuperscript{\rm 2}University of California, Santa Barbar\\
    \textsuperscript{\rm 3}University of California, Santa Cruz\\
    \textsuperscript{\rm 4}University of Electronic Science and Technology of China
}

\usepackage{bibentry}
\begin{document}

\maketitle

\begin{abstract}
Benchmarks are crucial for assessing multi-agent reinforcement learning (MARL) algorithms. While StarCraft II-related environments have driven significant advances in MARL, existing benchmarks like SMAC focus primarily on micromanagement, limiting comprehensive evaluation of high-level strategic intelligence. To address this, we introduce HLSMAC, a new cooperative MARL benchmark with 12 carefully designed StarCraft II scenarios based on classical stratagems from the \textit{Thirty-Six Stratagems}. Each scenario corresponds to a specific stratagem and is designed to challenge agents with diverse strategic elements, including tactical maneuvering, timing coordination, and deception, thereby opening up avenues for evaluating high-level strategic decision-making capabilities. We also propose novel metrics across multiple dimensions beyond conventional win rate, such as ability utilization and advancement efficiency, to assess agents' overall performance within the HLSMAC environment. We integrate state-of-the-art MARL algorithms and LLM-based agents with our benchmark and conduct comprehensive experiments. The results demonstrate that HLSMAC serves as a robust testbed for advancing multi-agent strategic decision-making.
\end{abstract}


\section{Introduction}
Multi-agent reinforcement learning (MARL) has achieved significant advancement, driven by diverse benchmarks across cooperative, competitive, and mixed settings. Environments like the StarCraft II Learning Environment (SC2LE \cite{vinyals2017starcraftiinewchallenge}) and its popular derivative, the StarCraft Multi-Agent Challenge (SMAC \cite{samvelyan19smac}) series, alongside other prominent testbeds, establish foundational standards that have fostered sophisticated algorithmic development.

However, current MARL benchmarks have several critical limitations that hinder progress in the field. The primary issue is that most existing benchmarks, including SMAC, emphasize micromanagement over strategic decision-making. Additionally, benchmark development presents a bottleneck because implementing robust testing environments is a time-consuming process. Furthermore, current benchmarks inadequately leverage established human strategic wisdom, as they primarily rely on emergent learning from environmental interaction, thus failing to assess agents' capacity for integrating human knowledge. 

Beyond traditional MARL approaches, the emergence of large language models (LLMs) provides a promising alternative paradigm for multi-agent decision-making. LLMs offer unique advantages through their advanced reasoning capabilities, interpretability, and inherent knowledge of human strategic principles. Recent research demonstrates their potential across diverse strategic environments, from simulating complex social interactions in Werewolf \cite{jin2024learning} to strategic planning in games such as Diplomacy \cite{meta2022human} and Chess \cite{feng2023chessgpt}. However, LLM-based approaches face challenges. The inherent issues such as hallucination and coordination complexities in multi-agent settings pose significant obstacles to developing robust multi-agent strategic intelligence.

To address these limitations, we introduce \textbf{HLSMAC} (\textbf{S}tarCraft \textbf{M}ulti-\textbf{A}gent \textbf{C}hallenge for \textbf{H}igh-\textbf{L}evel Strategic Decision-Making), a novel benchmark that translates classical Chinese Thirty-Six Stratagems into AI evaluation scenarios. We select 12 representative stratagems, each chosen for its clear principles and practical feasibility within game environments. Each stratagem is embodied in a dedicated StarCraft II map named after the stratagem. These maps are designed to challenge agents with diverse strategic elements, such as tactical maneuvering, timing coordination, and deception. The key innovations of HLSMAC include: (1) the systematic integration of established human strategic wisdom; (2) its emphasis on high-level strategic decision-making over micromanagement; and (3) compatibility with popular frameworks like PyMARL \cite{samvelyan19smac} and LLM-PySC2 \cite{li2024llm}.

Our contributions are as follows. First, we introduce HLSMAC, to our knowledge the first benchmark that systematically integrates the Thirty-Six Stratagems into multi-agent AI evaluation, shifting focus from micromanagement to high-level strategic decision-making. Second, we provide an extensive evaluation of 21 state-of-the-art MARL algorithms and the LLM-PySC2 framework, revealing current limitations and establishing a robust testbed. Third, we propose novel evaluation metrics beyond win rate to comprehensively assess strategic intelligence.

In this work, to ensure terminological clarity, we distinguish between three key concepts. A \textit{stratagem} refers to one of the classical Chinese Thirty-Six Stratagems. \textit{Macromanagement} refers to high-level strategic considerations such as economy and resource management, as distinguished from fine-grained unit control or micromanagement \cite{samvelyan19smac}. We define \textit{high-level strategic decision-making} as complex, human-like reasoning processes that involve formulating and executing overarching plans. Then we named our benchmark HLSMAC to emphasize both the high-level cognitive demands required for the tasks and the human-like strategic wisdom inspired by the Thirty-Six Stratagems.

\section{Related Work}
\subsection{Multi-agent Reinforcement Learning Benchmarks}

\subsubsection{StarCraft II-based Environments}
StarCraft II, as a real-time strategy game, provides an ideal testbed for multi-agent reinforcement learning with unique challenges including imperfect information, vast combinatorial action spaces, and long-term temporal credit assignment. Early works like TorchCraft \cite{synnaeve2016torchcraft} and SC2LE \cite{vinyals2017starcraftiinewchallenge} enable training on StarCraft games. Subsequently, SMAC \cite{samvelyan19smac} becomes one of the most popular benchmarks for cooperative multi-agent RL, focusing on decentralized micromanagement challenges. SMACv2 \cite{ellis2023smacv2} introduces procedurally generated scenarios and extended partial observability challenges to ensure agents must learn genuine closed-loop policies that condition on observations. SMAC-Hard \cite{deng2024smachardenablingmixedopponent} addresses the critical limitation of existing benchmarks where algorithms exploit specific weaknesses in static opponents rather than learning robust strategies, introducing opponent strategy editing, randomized opponent selection, and black-box testing frameworks. AlphaStar \cite{astar} and TStarBot-X\cite{han2021tstarbotxopensourcedcomprehensivestudy} both achieve competitive performance in StarCraft II.

\subsubsection{Other Prominent Testbeds}
Beyond early grid-world environments \cite{lowe2017multi, leibo2017multi}, MARL research has increasingly adopted more comprehensive platforms. OpenSpiel \cite{lanctot2019openspiel} offers a collection of environments and algorithms for MARL, alongside search and planning in games. For human-AI coordination, Overcooked-AI \cite{carroll2019utility} provides a dedicated collaborative cooking environment. FACMAC \cite{peng2021facmac} adapts continuous control environments like MuJoCo for multi-agent settings, while PettingZoo \cite{terry2021pettingzoo} offers a broad library covering competitive, cooperative, and mixed-sum games, including classic Atari and board games. For physics-based simulations, Google Research Football \cite{Kurach2019GoogleRF} provides a 3D soccer simulator with various benchmark tasks. More recently, Honor of Kings Arena \cite{Wei2022HonorOK} presents a competitive RL environment based on Honor of Kings, introducing new generalization challenges. Collectively, these platforms underscore the field's rapid progression.

\subsection{Methods for Solving MARL Benchmarks}
\subsubsection{MARL Algorithms} 
Multi-agent reinforcement learning (MARL) algorithms can be broadly categorized into two main approaches: value-based and policy-based. For value-based methods, IQL \cite{tampuu2017multiagent} treats other agents as part of the environment. To address the non-stationarity problem, VDN \cite{su2020value}, QMIX \cite{rashid2020monotonic}, QTRAN \cite{son2019qtran}, QPLEX \cite{wang2020qplex}, and others propose various methods to decompose the global Q-function into individual agent Q-functions. Qatten \cite{yang2020qattengeneralframeworkcooperative} employs multi-head attention structures for more accurate value decomposition. WQMIX \cite{rashid2020weightedqmixexpandingmonotonic} introduces weighting mechanisms to overcome the monotonicity constraints of standard mixing methods. Additional methods include ROMA \cite{wang2020romamultiagentreinforcementlearning} and RESQ \cite{pina2022residualqnetworksvaluefunction}, which further enhance coordination capabilities through various approaches, such as role-based decomposition mechanisms and improved network architectures. For policy-based methods, MADDPG \cite{lowe2020multiagentactorcriticmixedcooperativecompetitive} extends DDPG to multi-agent settings, BicNet \cite{peng2017multiagentbidirectionallycoordinatednetsemergence} introduces bidirectional communication, and COMA \cite{foerster2018counterfactual} employs counterfactual reasoning. LICA \cite{zhou2020learningimplicitcreditassignment} presents a multi-agent actor-critic method that formulates the centralized critic as a hypernetwork and employs adaptive entropy regularization. Additionally, trust region methods such as HATRPO and HAPPO \cite{kuba2021trust} provide theoretical guarantees for cooperative settings.

\subsubsection{LLM-based Methods} 
Large language models have shown potential in solving MARL problems. LLM-PySC2 \cite{li2025llmpysc2starcraftiilearning} introduces a novel environment that enables large language models to interact with StarCraft II. ChessGPT \cite{feng2023chessgpt} demonstrates LLM's potential to integrate strategic learning with language understanding in complex decision-making tasks. \cite{jin2024learning} emphasizes the importance of strategic discussion tactics in shaping player beliefs and game outcomes in Werewolf. The GITM \cite{zhu2023ghost} and Voyager \cite{wang2023voyager} projects based on the MineDojo environment showcase LLM agents' capabilities in navigation and task execution within Minecraft. Recent work, such as \cite{chen2024vlmsplayactionroleplaying}, demonstrates LLM agents' ability to master complex visual-based combat mechanics in RPGs.

\section{HLSMAC}
In this section, we firstly examine the key design features considered in constructing HLSMAC scenarios to support high-level strategic decision-making, and then illustrate how human strategic wisdom is incorporated into the scenarios via three examples.

\subsection{Design Features of Scenarios}

HLSMAC fully leverages the inherent characteristics of StarCraft II to enable the evaluation of high-level strategic decision-making over micromanagement. Specifically, we utilize the game's rich strategic complexity, flexible map editor, and diverse official map pool to create tailored HLSMAC scenarios. The following details the core design features of the scenarios.

\subsubsection{Larger Map Sizes and Richer Terrain Elements}
To enhance strategic complexity, each map features at least 80×80 grids, compared to SMAC's standard 32×32 layouts, providing longer movement distances, more route options, and additional combat zones. These expanded dimensions also make it feasible to place more units and structures, as well as design more diverse terrain, including high grounds, choke points, and open fields. Although creating such terrains is typically challenging, we streamline the process by cropping from official StarCraft II ladder maps, reproducing authentic battlefield dynamics. We expect agents to exploit spatial and terrain features for strategic decisions, rather than focusing solely on micromanagement.

\subsubsection{Expanded Unit and Structure Abilities}
We introduce more game-inherent abilities for units and structures to expand the action space. In contrast with SMAC, where unit abilities are restricted to basic functions like \texttt{move}, \texttt{stop}, and \texttt{attack} that primarily support micromanagement, HLSMAC's diverse abilities are highly relevant to the specific scenario objectives and can enhance task performance when properly utilized. For units, Zerglings can use \texttt{Burrow} to enable ambushes or retreats, while Sentries may cast \texttt{Hallucination} to create fake allies, deceiving opponents about the true force composition. For structures, operational abilities are selectively extended, such as the \texttt{Load} and \texttt{Unload} abilities for Nydus Worms, which facilitate rapid unit transportation, and the \texttt{WarpIn} ability for Warp Gates, enabling instant unit deployment. Considering PySC2 library compatibility for action space extension, we use built-in StarCraft II units (and structures) instead of SMAC's RL units. Such abilities indicate strategic thinking that involves deception, misdirection, and long-term planning, thereby shifting the action space from reactive to strategic.

\subsubsection{Diverse Opponent Policies}
We define diverse opponent policies by utilizing the built-in trigger system of the StarCraft II Editor. For example, some scenarios feature triggers that order enemy forces to attack proactively at game start to simulate aggressive opponents. In contrast, to emulate enemies retreating when facing a numerical disadvantage, a trigger activates when agent units entering the enemy's sight range exceed a specified threshold, immediately prompting all enemies to move back to their base. Through careful trigger design involving numerical calculations, logical operations, and conditional statements, these triggers enable realistic opponent responses, introducing tactical complexity.

\subsubsection{Redefined Game Termination Conditions}
We redefine game termination conditions for HLSMAC scenarios. Rather than using the elimination of all enemy units as the sole victory condition, we implement more diverse success criteria, such as destroying critical structures or maintaining unit survival for specified periods. For instance, as described in the Thirty-Six Stratagems, targeting key assets is often more effective than engaging in exhaustive confrontation. This shift from elimination-based victory conditions to strategic objectives encourages higher-level strategic planning rather than micro-level optimization.

We comprehensively integrate the above design features to carefully craft scenarios that require high-level strategic reasoning to succeed. We expect most existing methods to fail inevitably unless they follow the stratagem.

\subsection{Scenarios}
The Thirty-Six Stratagems is a unique and well-known collection of ancient Chinese proverbs that describe some of the most cunning and subtle tactics ever devised. This classical work represents a sophisticated codification of human strategic thinking. To our knowledge, HLSMAC is the first benchmark to systematically integrate the stratagems into the research on multi-agent intelligence.

\begin{figure}[t]
\centering
\includegraphics[width=1\columnwidth]{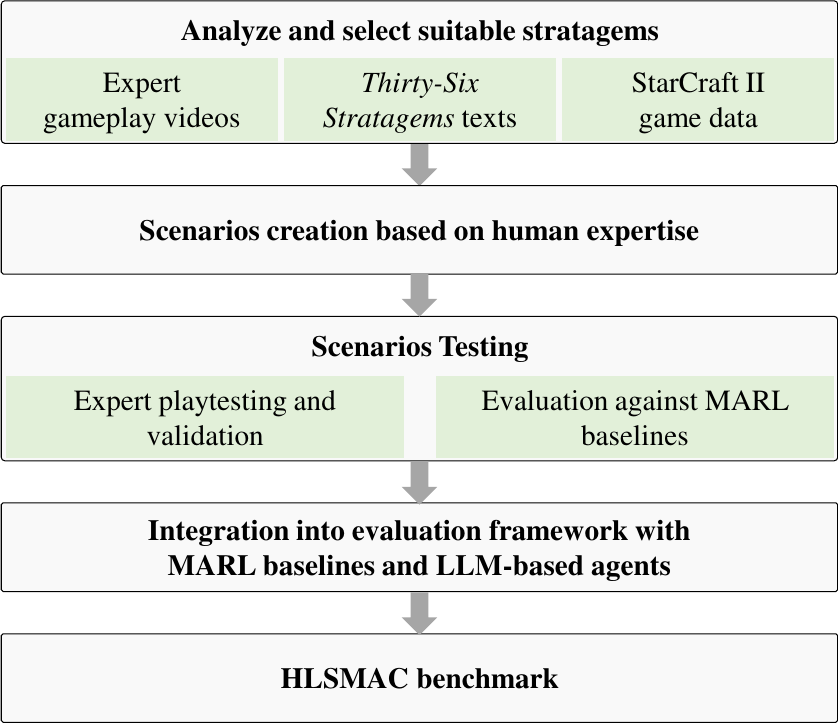} 
\caption{Benchmark Construction Pipeline}
\label{fig:pipline}
\end{figure}

We adopt the benchmark construction pipeline as shown in Figure~\ref{fig:pipline}. First, we comprehensively gather and study the Thirty-Six Stratagems texts, gameplay videos of human players applying these stratagems, and relevant StarCraft II game data. Following this, we analyze the core strategic concepts of these stratagems and select those most suitable for the StarCraft II environment. Next, we leverage human expertise to design scenario storylines, then implement them by configuring terrain, game mechanics, and unit abilities through StarCraft II Editor. Each scenario is iteratively refined through testing to ensure that it accurately captures the essence of the stratagem. 

We eventually select 12 representative stratagems, each chosen for its clear strategic principles and practical feasibility within game environments. The complete list of 12 HLSMAC scenarios is presented in Table~\ref{tab:table-hlsmac-scenarios}, which provides a brief overview of the unit compositions for each scenario. Maps are named using the initials of the stratagems' Chinese names. The stratagems and detailed scenario descriptions for HLSMAC are provided in Appendices A.1 and A.2.

\begin{table*}[t]
\centering
{\fontsize{10pt}{\baselineskip}\selectfont
\begin{tabular}{ccc}
\hline
\textbf{Name} & \textbf{Ally Units and   Structures}    & \textbf{Enemy Units and   Structures}   \\ \hline
\textbf{adcc} & 16 Zerglings, 1 Hatchery     & 4   Hellbats, 1 Command Center    \\ \hline
\shortstack{\textbf{dhls}\\~} & \shortstack{9 Zerglings, 4 Roaches, 1 Hatchery,\\ 1 Nydus Network, 1 Nydus Worm} & \shortstack{8 Marines, 2 Siege Tanks, 1 Command Center\\~} \\ \hline
\textbf{fkwz} & 2 Warp Gates, 2 Pylons, 1 Warp Prism     & 1   Stalker, 1 Gateway, 2 Pylons, 3 Photon Cannons      \\ \hline
\textbf{gmzz} & 5 Marines, 3 Supply Depots     & 8   Zerglings, 3 Spine Crawlers          \\ \hline
\textbf{jctq} & 4 Roaches     & 2   Sentries, 6 Stalkers, 1 Observer         \\ \hline
\textbf{jdsr} & 4 Roaches, 1 Infestor & 3   Stalkers, 1 Colossus         \\\hline
\textbf{sdjx} & 14 Marines, 4 Medivacs   & 5   Zealots, 4 Stalkers, 3 Colossi, 2 Nexuses, 2 Assimilators, 1 Pylon \\\hline
\textbf{swct} & 4 Sentries, 1 Warp Prism         & 10   Zerglings, 1 Hatchery                                             \\ \hline
\textbf{tlhz} & 1 Drone, 3 Larvas & 1   Photon Cannon, 1 Pylon                                             \\\hline
\textbf{wwjz} & 7 Zealots, 1 Nexus          & 8   Hellions, 6 Marines, 1 Command Center                              \\\hline
\textbf{wzsy} & 2 Stalkers, 4 Sentries           & 3   Immortals, 5 Zealots, 1 Nexus, 1 Pylon                             \\\hline
\textbf{yqgz} & 24 Zerglings  & 6   Marines, 2 Siege Tanks            \\ \hline
\end{tabular}}
\caption{HLSMAC Scenarios}
\label{tab:table-hlsmac-scenarios}
\end{table*}

\begin{figure*}[t]
\centering
\includegraphics[width=\textwidth]{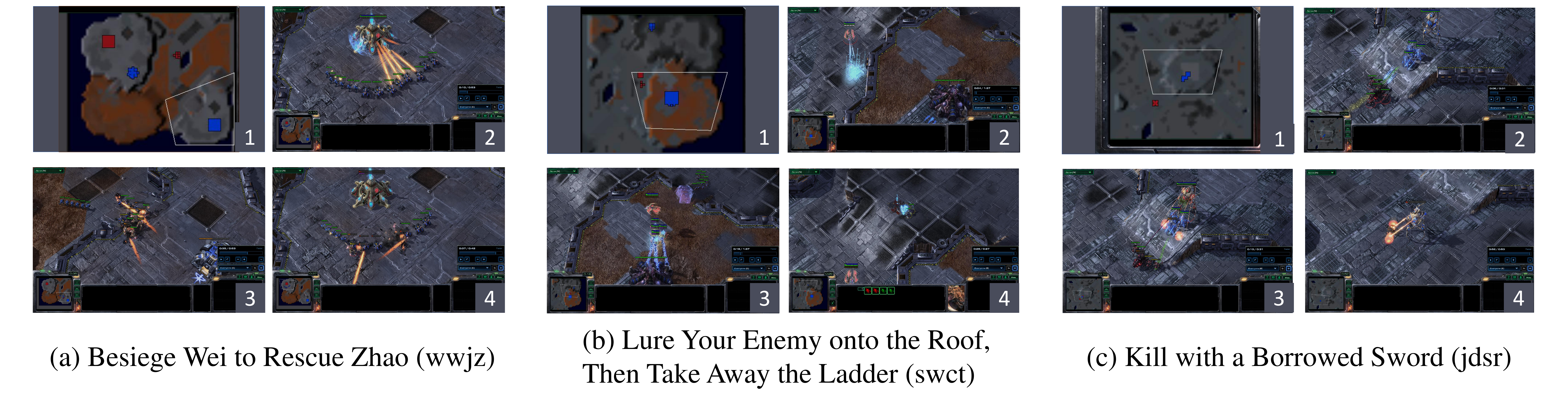}
\caption{Scene 1 of each subfigure shows the initial battlefield (red: our forces, blue: enemy). Scenes 2-4 depict the early stage, the victory stage (with stratagem), and the defeat stage (without stratagem), corresponding to each scenario. As in these three examples, the stratagem's use or absence yields opposite outcomes across all HLSMAC scenarios.}
\label{fig:scenario_examples}
\end{figure*}

We now present three representative examples to illustrate the integration of human strategic wisdom into HLSMAC scenarios. For each example, we examine the original stratagem, describe the scenario mechanics, and propose the expected solution (see Figure~\ref{fig:scenario_examples}).

\subsubsection{Example 1: Besiege Wei to Rescue Zhao (wwjz)}
This stratagem derives its name from a famous ancient incident that occurred in 354 B.C. When Wei forces besieged Zhao's capital, to rescue Zhao from its predicament, the strategists chose not to directly attack the besieging army but instead to attack Wei's own capital, forcing Wei to abandon its siege and rush back to defend its homeland. The exhausted Wei troops were then ambushed and defeated during their retreat. Its strategic logic is articulated as follows: When the enemy is too strong to attack directly, strike at something he holds dear.

Based on this stratagem, we design the following scenario. When the game starts, our Nexus (upper-left) is under frantic assault by 8 Hellions and 6 Marines, with no local defenses. Meanwhile, our 7 Zealots are positioned in the upper-right region of the map. Our objective is clear: either destroy the enemy Command Center or eliminate all enemy units, all while ensuring our Nexus survives. To simulate the core stratagem, a crucial trigger is in place: the moment our Zealots approach the enemy Command Center, all enemy forces will immediately disengage from our Nexus and retreat to defend their own base, mirroring the dynamics of \textit{Besiege Wei to Rescue Zhao}. Through extensive testing, we have calibrated the scenario to ensure that our Zealots cannot reach the Nexus in time for a direct defense. Even if the Zealots retreat and engage the enemy directly, we are likely to lose.

Correspondingly, the expected solution is as follows. Rather than engaging the superior enemy force in direct combat, the Zealots should encircle the enemy Command Center, embodying the \textit{Besiege Wei} principle. This forces the Hellions and Marines to abandon their assault on the Nexus and retreat to defend their base, thereby achieving the \textit{Rescue Zhao} effect. Furthermore, due to the speed difference between Hellions and Marines, the enemy forces return in a scattered formation, allowing the Zealots to defeat them in successive waves as they arrive piecemeal at their base.

% \begin{figure}[t]
% \centering
% \includegraphics[width=1\columnwidth]{swct.pdf} % Reduce the figure size so that it is slightly narrower than the column. Don't use precise values for figure width.This setup will avoid overfull boxes.
% \caption{Lure Your Enemy onto the Roof, Then Take Away the Ladder (swct)}
% \label{swct}
% \end{figure}

\subsubsection{Example 2: Lure Your Enemy onto the Roof, Then Take Away the Ladder (swct)}
This stratagem derives its core principle from Sun Tzu's brilliant deception at the Battle of Maling in 341 B.C. When facing a powerful enemy's pursuit, the stratagem avoids direct engagement, but instead designs an elaborate trap to lure the enemies deeper, subsequently blocking their line of advance and destroying the entire force in a narrow valley.

Inspired by this stratagem, we design the following scenario. When the game starts, our 1 Warp Prism (with \texttt{Load} and \texttt{Unload} abilities) and 4 Sentries (capable of casting \texttt{Force Field}) face a complex challenge. The enemy Hatchery is located on nearby high ground, accessible only through a narrow choke, and along the path to this choke, 10 enemy Zerglings are stationed as defenders. Our goal is either to destroy the enemy Hatchery or eliminate all the Zerglings. The trigger is configured as follows: once our forces approach the Hatchery, all enemy Zerglings will immediately retreat to defend their own base. It must be emphasized that due to the disparity in unit numbers, our forces cannot achieve victory through direct engagement with the Zerglings.

Similarly, the expected solution requires human-like strategic thinking. The Warp Prism should bypass the enemy Zerglings' frontline and deploy Sentries onto the high ground. Once the enemy Hatchery is under a surprise attack, the Zerglings will retreat, allowing the Sentries to maintain Force Field barriers at the choke point. This either delays reinforcements to destroy the Hatchery or isolates and eliminates returning Zerglings, thus mirroring and executing this classic stratagem.

\subsubsection{Example 3: Kill with a Borrowed Sword (jdsr)}
This stratagem embodies a fundamental principle: when lacking sufficient strength, harness the enemy's power for your purposes. If direct confrontation proves impossible, turn the enemy's weapons against them.

Drawing from this stratagem, we develop the following scenario. At game start, our 4 Roaches and 1 Infestor, which can use \texttt{Neural Parasite} to temporarily take control of target units, are positioned in the center area, preparing for imminent combat as the enemy's 3 Stalkers and 1 Colossus advance from the upper right via the ramp. The trigger mechanism is straightforward: enemy forces initiate combat at scenario onset. Our objective is to eliminate all enemy units. Given the Colossus's superior firepower and mobility, direct engagement would result in defeat.

Success requires turning the enemy's strength against itself. The Infestor immediately casts \texttt{Neural Parasite} to control the enemy Colossus, turning it against the Stalkers as all forces rush to eliminate every enemy unit.

\section{Evaluation Frameworks}
From the outset, the HLSMAC benchmark aims to interface with both Multi-Agent Reinforcement Learning (MARL) and Large Language Model (LLM) frameworks, providing a unified testbed that enables the exploration of how different AI paradigms tackle high-level strategic decision-making tasks. To achieve this, our implementation specifically integrates PyMARL for MARL approaches and LLM-PySC2 for large language model applications.

Both frameworks share a common foundation that enables this integrative capability: the StarCraft II interface ecosystem, developed by Blizzard, DeepMind, and the research community, comprising StarCraft II binaries (for Windows and Linux), the StarCraft II API, and PySC2. Taking PySC2 as a key example, this open-source Python library provides general capabilities that extend beyond its initial RL optimization, enabling diverse agents to extract observations, execute actions, and access game state information for strategic intelligence development.

Despite sharing a common foundation, the two frameworks serve distinct purposes. PyMARL focuses on training reinforcement learning models and supports multiple environments, whereas LLM-PySC2 specializes in large language model inference for StarCraft II scenarios. Below we present the integration details for both frameworks.

\subsection{PyMARL Framework}
PyMARL\cite{samvelyan19smac}, initially built for developing MARL algorithms for SMAC, has become a widely adopted, open-source framework across diverse multi-agent environments. Its modular architecture facilitates the integration of state-of-the-art algorithms. Numerous fork implementations with various baselines have emerged, providing an ideal evaluation platform for HLSMAC. 

It is worth noting that PyMARL serves solely as a training framework, decoupled from environment components. To integrate HLSMAC with PyMARL's baseline algorithms, our core effort focuses on developing a dedicated environment codebase for its game maps. 

\subsubsection{HLSMAC Environment Implementation}
HLSMAC environment uses the factory pattern for modular design. A shared BaseEnv class handles common logic, while each of the 12 scenarios inherits from it to implement specific action spaces, unit state updates, and game termination conditions. This design achieves logical separation of scenarios, promotes code reusability, and facilitates extensibility for new scenarios. Additionally, by mirroring SMAC's environment wrapper design, HLSMAC enables backward compatibility with existing SMAC maps.

Considering that certain abilities in HLSMAC scenarios, such as \texttt{Hallucination} and \texttt{WarpIn} create new units mid-game, our environment implements support for dynamic unit spawning, a feature unavailable in SMAC. To handle this dynamism, the environment pre-allocates unit slots at the beginning of each episode, initially set to None for potential spawning. These empty slots are padded with zero-filled observation and state vectors, along with \texttt{no-op} actions. When units spawn, the corresponding None slots are replaced with actual unit data, enabling flexible scalability for complex multi-agent scenarios.

HLSMAC scenarios feature diverse victory conditions, such as destroying critical structures or maintaining unit survival for specified periods. Given this diversity, HLSMAC implements tailored termination logic for each scenario. Additionally, the environment enhances the reward tracking mechanism by independently logging reward components, including delta\_ally, delta\_enemy, and delta\_death, to help analyze agent training processes.

HLSMAC adopts the same data structures for state and observation as SMAC, primarily for maintaining compatibility with existing baselines.

\subsubsection{Interface with PyMARL}
In practice, interfacing the HLSMAC environment with existing PyMARL frameworks requires minimal modification, typically just registering the new environment and updating the environment selector logic with a few lines of code. This simplicity stems from HLSMAC's close alignment with the underlying implementation of the SMAC environment, enabling straightforward algorithm integration and evaluation within PyMARL's framework.

\subsection{LLM-PySC2 Framework}
\subsubsection{Extend LLM Agents Configuration}
The LLM-PySC2 framework employs a hierarchical agent system comprising a MainAgent and SubAgents. MainAgent coordinates the SubAgents, while each SubAgent interfaces with the LLM to handle specific tasks like building construction and unit production.

We extend the framework from Protoss-only to all three races by configuring additional agents for HLSMAC scenarios. The extension process follows two core principles. The first is designing agents based on specific scenario requirements. For example, in the ``Shut the Door to Catch the Thief'' scenario, we develop a dedicated Supply Depot agent to handle the complexity of depot management. The second approach is to group the agents by unit attributes or functional similarity. In the ``Clamour in the East, Attack in the West'' scenario, Medivacs join the Air Force group while Marines join the Ground Force group. This grouping mirrors human gameplay logic while maximizing framework modularity.

\subsubsection{Interface with LLM Agents}
To support HLSMAC, we implement an agent interface that enables scenario selection and task prompt configuration, driving agents to decompose the task into action sequences to succeed. Through this approach, we can examine whether agents can understand abstract semantics by primarily using explanations of specific stratagems as task prompts. The framework also supports more detailed prompts for further study.

\section{Evaluation Metrics and Results}
As a real-time strategy game, StarCraft II provides comprehensive performance indicators, including average unspent resources, time supply capped, workers created, and APM (Actions Per Minute). For instance, APM is often used to measure a player's operational speed and competitive level. These metrics target comprehensive, match-long performance based on the full map and are not suitable for HLSMAC scenarios. Therefore, we carefully examine the existing framework-related metrics and introduce HLSMAC's scenario-specific metrics.

\subsection{Exsiting Framework-Related Metrics}
Most of the MARL algorithms offer two types of evaluation metrics. Training performance metrics track learning dynamics and convergence, encompassing optimization indicators such as loss, td\_error\_abs, and target\_mean, reward statistics including reward\_mean and reward\_std, and combat effectiveness measures like battle\_won\_mean, dead\_allies\_mean, and dead\_enemies\_mean. On the other hand, test performance metrics utilize corresponding test versions of training metrics prefixed with ``test\_''. The test\_battle\_won\_mean serves as the primary win-rate indicator. This dual-metric framework enables comprehensive assessment of both algorithmic learning capabilities and practical combat effectiveness.

Similar to MARL methods, LLM-PySC2 evaluates the capabilities of large models based on winning\_rates, kill\_rates, and death\_rates. 

\subsection{HLSMAC Scenario-Specific Metrics}

We propose a set of metrics broadly applicable across diverse baselines within HLSMAC scenarios. These metrics assess performance from several additional dimensions, including critical target advancement, ability utilization frequency, target damage, and unit survival rate. To calculate these metrics, we extract data from the game replay files generated by the evaluated algorithms. 

Each metric is defined as follows, where $N$ is the number of game episodes for each scenario (map).

\texttt{Critical\string_Target\string_Advancement} measures the movement of allied units toward enemy critical targets. These targets typically refer to main structures such as Command Center and Hatchery, strategic coordinates, and key enemy units, varying by scenario. This metric can be computed through the following two algorithms.

\begin{itemize}
    \item Target Proximity Frequency (TPF) reflects the average frequency of allied units entering within a specified distance range around enemy critical targets per game episode.
    \begin{equation}
        \mathrm{TPF} =  \frac{1}{NM}\sum_{i=1}^{N}\sum_{j=1}^{M} \mathbf{1}(d_{ij} \leq L)
    \end{equation}
    Where $M$ is the number of allied combat units in the scenario, $d_{ij}$ is the distance between the $j$-th unit and the enemy critical targets at any game loop in episode $i$, $L$ is a threshold defining the distance range around the target (e.g., $L = 6$), and $\mathbf{1}(\cdot)$ is the indicator function.
    
    \item Target Directional Alignment (TDA) computes the projection ratio of actual displacement vectors onto shortcut vectors.
    \begin{equation}
        \mathrm{TDA} = \frac{1}{NM} \sum_{i=1}^{N} \sum_{j=1}^{M} \frac{\vec{v}_{ij} \cdot \vec{t}_{ij}}{||\vec{t}_{ij}||^2}
    \end{equation}
    Where $\vec{v}_{ij}$ is the actual displacement vector of the $j$-th unit in episode $i$ (final position - initial position), $\vec{t}_{ij}$ is the shortcut vector from the initial position to the enemy critical targets. Thus, we can measure the effectiveness of the movement towards the target.
\end{itemize}

\texttt{Ability\string_Utilization\string_Frequency} measures how frequently allied units cast special abilities during gameplay. Such abilities include \texttt{Burrow}/\texttt{Unburrow}, \texttt{Load}/\texttt{Unload}, and other actions beyond basic movement and attack.
\begin{equation}
    \mathrm{AUF} = \frac{1}{N}\sum_{i=1}^{N}\sum_{k=1}^{K} A_{ik}
\end{equation}

Where $K$ is the number of distinct ability types tracked, and $A_{ik}$ is the count of type $k$ ability cast in episode $i$.

\texttt{Critical\string_Target\string_Damage} measures the total damage dealt by allied units to enemy critical targets during gameplay.

\begin{equation}
   \mathrm{CTD} = \frac{1}{N}\sum_{i=1}^{N}\sum_{j=1}^{J} \frac{D_{ij}}{D_{max,j}}
\end{equation}

Where $J$ is the number of enemy critical target types, $D_{ij}$ is the total damage dealt to type $j$ enemy critical targets in episode $i$, and $D_{max,j}$ is the total health points of type $j$ enemy critical targets.

\texttt{Unit\string_Survival\string_Rate} is the average ratio of surviving units to initial unit count. It measures the AI's capability of preserving units during gameplay.

\begin{equation}
    \mathrm{USR} = \frac{1}{N}\sum_{i=1}^{N} \frac{U_{remaining,i}}{U_{initial,i}}
\end{equation}

Where $U_{remaining,i}$ is the number of allied units surviving at the end of episode $i$, and $U_{initial,i}$ is the initial number of allied units at the beginning of episode $i$.

\subsection{Results}
We evaluate 21 MARL algorithms from open-source repositories using default parameters, along with GPT-3.5 agents through LLM-PySC2. The experimental configuration and complete results are provided in Appendices A.3 and A.4. Below, we identify three key findings.

Firstly, HLSMAC poses significant challenges for both MARL and LLM-based methods. Nearly $80\%$ of algorithm-scenario combinations achieve zero win rates as shown in Table~\ref{tab:marl_performance}. Note that scenarios where all algorithms achieved zero win rates are excluded from the table presentation. Furthermore, ChatGPT 3.5-based agents fail across all scenarios despite showing limited strategic understanding in HLSMAC scenarios.

Secondly, traditional win-rate metrics inadequately capture agents' human-like strategic decision-making capabilities in HLSMAC tasks. For example, RIIT achieves a $93\%$ win rate in \textit{adcc}, but replay analysis reveals it does not actually follow the intended stratagem approach. Similarly, DOP achieves only a $19\%$ win rate in \textit{swct}, yet exhibits repeated loading and unloading of Sentries with Warp Prism, a behavior rarely seen in normal human play. In addition, high win rates in \textit{jdsr} do not guarantee that the Infester actually comprehends the ``borrow sword'' concept, even when casting \texttt{Neural Parasite}.

Finally, our new metrics, together with win rate, offer richer dimensions for evaluating the performance of various methods. Through $R^2$ analysis of valid metric-scenario combinations, we find that these metrics demonstrate strong explanatory power ($R^2\geq 0.6$) in 44.4\% of the combinations. Metrics such as TPF, TDA, and CTD serve as effective indicators of high win-rate performance in the \textit{wzsy}, \textit{wwjz}, and \textit{sdjx} scenarios. This reflects a fundamental strategic truth: success depends on approaching and neutralizing critical targets, which the \textit{wwjz} and \textit{sdjx} scenarios clearly exemplify. Furthermore, although AUF cannot directly reflect the appropriateness of ability usage, it can distinguish between human players and current methods, as humans demonstrate purposeful utilization of critical abilities, whereas current methods do not. For instance, in the \textit{gmzz} scenario, human experts use \texttt{SupplyDepotLower} and \texttt{SupplyDepotRaise} only once each to strategically block and contain the enemy Zerglings. Current MARL methods, however, usually execute these abilities hundreds of times without clear intent.

\begin{table}[t] % [t] places the table at the top of the page
\centering
\setlength{\tabcolsep}{1mm}
{\fontsize{9pt}{\baselineskip}\selectfont
% \begin{table}[]
% \resizebox{\columnwidth}{!}{%
\begin{tabular}{|c|c|c|c|c|c|c|c|c|}
\hline
       & \textbf{adcc} & \textbf{gmzz} & \textbf{jctq} & \textbf{jdsr} & \textbf{sdjx} & \textbf{swct} & \textbf{wwjz} & \textbf{wzsy} \\ \hline
IQL    & 0.3 & 0.04 &0& 0.4 &0& 0.03 &0& 0.34 \\ \hline
COMA   &0&0&0& 0.02 &0&0&0&0 \\ \hline
VDN    &0& 0.12 &0& 0.6 & 0.14 &0&0& 0.09 \\ \hline
QMIX   &0& 0.01 &0& 0.88 &0&0&0& 0.85 \\\hline
VMIX   &0&0&0&0&0&0&0&0 \\\hline
MAVEN  &0&0&0& 0.74 &0&0&0&0 \\\hline
QTRAN  &0& 0.12 &0& 0.82 & 0.32 &0&0&0 \\\hline
CWQMIX &0&0&0& 0.42 & 0.83 &0& 0.01 & 0.96 \\\hline
OWQMIX &0&0&0& 0.85 &0&0&0&0 \\\hline
DOP    &0& 0.26 &0&1&0& 0.19 &0&0 \\\hline
LICA   &0& 0.50 &0& 0.73 &0&0&0& 0.02 \\\hline
Qatten &0& 0.15 &0& 0.92 & 0.88 &0&0& 0.94 \\\hline
QPLEX  &0&0&0& 0.93 & 0.86 &0&0&0 \\\hline
FOP    &0&0& 0.03 & 0.73 &0&0&0&0 \\\hline
RIIT   & 0.93 & 0.27 &0& 0.96 &0&0&0&0 \\\hline
RODE   &0&0&0&0&0&0&0&0 \\\hline
ROMA   &0&0&0& 0.77 &0&0&0&0 \\\hline
RESQ   &0& 0.41 &0& 0.91 &0&0&0&0 \\\hline
RESZ   &0& 0.26 &0& 0.79 & 0.72 &0& 0.78 & 0.07 \\\hline
dTAPE  & 0.49 & 0.84 &0& 0.93 & 0.89 &0&1& 1 \\\hline
sTAPE  &0&0&0& 0.98 &0&0&0& 0 \\ \hline
\end{tabular}%
}
\caption{Win Rates Across 21 MARL Baselines on Non-zero Scenarios (with \textit{dhls}, \textit{fkwz}, \textit{tlhz}, \textit{yqgz} excluded)}
\label{tab:marl_performance}
\end{table}

\section{Conclusion}
In this paper, we propose HLSMAC, a new StarCraft multi-agent challenge for high-level strategic decision-making, the first environment that systematically integrates classical Chinese Thirty-Six Stratagems into AI evaluation scenarios. HLSMAC focuses on cooperative multi-agent decision-making and supports frameworks like PyMARL and LLM-PySC2 for evaluation. We introduce richer metrics beyond conventional win rates to assess agents' strategic performance within the HLSMAC environment. We conduct comprehensive experiments, and the results demonstrate that HLSMAC serves as a robust testbed for advancing multi-agent research. While our current work establishes the foundation for strategic evaluation, several promising directions remain for future exploration, including the development of automated scenario generation methodologies and specialized algorithms tailored to HLSMAC's unique challenges.

\bibliography{aaai2026}

% If so, you can uncomment the following line and adjust the path to include it.
%\input{../../ReproducibilityChecklist/LaTeX/ReproducibilityChecklist.tex}
% %File: anonymous-submission-latex-2026.tex
% \documentclass[letterpaper]{article} % DO NOT CHANGE THIS
% \usepackage[submission]{aaai2026}  % DO NOT CHANGE THIS
% \usepackage{times}  % DO NOT CHANGE THIS
% \usepackage{helvet}  % DO NOT CHANGE THIS
% \usepackage{courier}  % DO NOT CHANGE THIS
% \usepackage[hyphens]{url}  % DO NOT CHANGE THIS
% \usepackage{graphicx} % DO NOT CHANGE THIS
% \urlstyle{rm} % DO NOT CHANGE THIS
% \def\UrlFont{\rm}  % DO NOT CHANGE THIS
% \usepackage{natbib}  % DO NOT CHANGE THIS AND DO NOT ADD ANY OPTIONS TO IT
% \usepackage{caption} % DO NOT CHANGE THIS AND DO NOT ADD ANY OPTIONS TO IT
% \frenchspacing  % DO NOT CHANGE THIS
% \setlength{\pdfpagewidth}{8.5in} % DO NOT CHANGE THIS
% \setlength{\pdfpageheight}{11in} % DO NOT CHANGE THIS
% %
% % These are recommended to typeset algorithms but not required. See the subsubsection on algorithms. Remove them if you don't have algorithms in your paper.
% \usepackage{algorithm}
% \usepackage{algorithmic}

% %
% % These are are recommended to typeset listings but not required. See the subsubsection on listing. Remove this block if you don't have listings in your paper.
% \usepackage{newfloat}
% \usepackage{listings}
% \DeclareCaptionStyle{ruled}{labelfont=normalfont,labelsep=colon,strut=off} % DO NOT CHANGE THIS
\lstset{%
	basicstyle={\footnotesize\ttfamily},% footnotesize acceptable for monospace
	numbers=left,numberstyle=\footnotesize,xleftmargin=2em,% show line numbers, remove this entire line if you don't want the numbers.
	aboveskip=0pt,belowskip=0pt,%
	showstringspaces=false,tabsize=2,breaklines=true}
\floatstyle{ruled}
\newfloat{listing}{tb}{lst}{}
\floatname{listing}{Listing}
%
% Keep the \pdfinfo as shown here. There's no need
% for you to add the /Title and /Author tags.
\pdfinfo{
/TemplateVersion (2026.1)
}

\setcounter{secnumdepth}{2} %May be changed to 1 or 2 if section numbers are desired.

% The file aaai2026.sty is the style file for AAAI Press

% Title

% Your title must be in mixed case, not sentence case.
% That means all verbs (including short verbs like be, is, using,and go),
% nouns, adverbs, adjectives should be capitalized, including both words in hyphenated terms, while
% articles, conjunctions, and prepositions are lower case unless they
% directly follow a colon or long dash
\title{HLSMAC: A New StarCraft Multi-Agent Challenge for High-Level Strategic Decision-Making}
\author{
    %Authors
    % All authors must be in the same font size and format.
    % Written by AAAI Press Staff\textsuperscript{\rm 1}\thanks{With help from the AAAI Publications Committee.}\\
    % AAAI Style Contributions by Pater Patel Schneider,
    % Sunil Issar,\\
    % J. Scott Penberthy,
    % George Ferguson,
    % Hans Guesgen,
    % Francisco Cruz\equalcontrib,
    % Marc Pujol-Gonzalez\equalcontrib
}

\maketitle

\appendix
\section{Appendix}
\subsection{HLSMAC Stratagems}
The Thirty-Six Stratagems is a renowned collection of ancient Chinese strategic proverbs describing sophisticated tactical principles. We have carefully selected 12 representative stratagems for their clear principles and practical feasibility within the HLSMAC environment. Each stratagem is embodied in a dedicated StarCraft II map, where core strategic concepts are integrated into specific game scenarios. It is worth noting that researchers can leverage these core concepts to design their own novel scenarios.

Each entry below includes the stratagem name, Chinese pinyin pronunciation, explanation, and the key connection to the corresponding scenario. Additional details can be found in the English translations of the Thirty-Six Stratagems (e.g., Verstappen's \textit{The Thirty-Six Strategies of Ancient China}) and online resources such as Wikipedia.

\begin{enumerate}
\item \textbf{Stratagem: Secretly March to Chencang (adcc)}
\begin{itemize}
\item Pinyin: An du chen cang
\item Explanation: Use an unexpected indirect attack when the enemy is preparing his defense on your obvious direct attack. This will cause the enemy to divide his forces at the last minute, leading to confusion and disaster.
\item Connection to HLSMAC scenario: Use Zergling's burrow ability to launch an unexpected indirect attack on the enemy's base.
\end{itemize}
\item \textbf{Stratagem: Lure the Tiger Down the Mountain (dhls)}
\begin{itemize}
\item Pinyin: Diao hu li shan
\item Explanation: Never directly attack a well-entrenched opponent. Instead, lure him away from his stronghold and separate him from his source of strength.
\item Connection to HLSMAC scenario: Use small allied forces to lure strong defensive enemy units away from their base, enabling a sneaky base assault.
\end{itemize}
\item \textbf{Stratagem: Exchange the Role of Guest for That of Host (fkwz)}
\begin{itemize}
\item Pinyin: Fan ke wei zhu
\item Explanation: Defeat the enemy by infiltrating the enemy's camp under the guise of cooperation, surrender, or peace treaties. In this way, you can discover his weakness and then, when the enemy's guard is relaxed, strike directly at the source of his strength.
\item Connection to HLSMAC scenario: Use Warp Prism's mobility to infiltrate the enemy base and warp in Zealots directly at their production facilities, striking at the source of their strength.
\end{itemize}
\item \textbf{Stratagem: Shut the Door to Catch the Thief (gmzz)}
\begin{itemize}
\item Pinyin: Guan men zhuo zei
\item Explanation: If you have the chance to completely capture the enemy, then you should do so, thereby bringing the battle or war to a quick and lasting conclusion. Do not allow your enemy to escape, it can plant the seeds for future conflict. However, if they succeed in escaping, be wary of giving chase.
\item Connection to HLSMAC scenario: Use Supply Depot elevation control to lure enemy forces, then raise the Depots to trap them and prevent their escape.
\end{itemize}
\item \textbf{Stratagem: Shed Your Skin Like the Golden Cicada (jctq)}
\begin{itemize}
\item Pinyin: Jin chan tuo qiao
\item Explanation: When you are in danger of being defeated, and your only chance is to escape and regroup, then create an illusion. While the enemy's attention is focused on this artifice, secretly remove your men, leaving behind only the facade of your presence.

\item Connection to HLSMAC scenario: Use Roach's ability to move while burrowed to escape under attack.
\end{itemize}
\item \textbf{Stratagem: Kill with a Borrowed Sword (jdsr)}
\begin{itemize}
\item Pinyin: Jie dao sha ren
\item Explanation: When lacking sufficient strength, harness the enemy's power for your purposes. If direct confrontation proves impossible, turn the enemy's weapons against them.
\item Connection to HLSMAC scenario: Use Infestor's Neural Parasite to control the enemy's most powerful unit to against themselves.
\end{itemize}

\item \textbf{Stratagem: Clamour in the East, Attack in the West (sdjx)}
\begin{itemize}
\item Pinyin: Sheng dong ji xi
\item Explanation: In any battle, the element of surprise can provide an overwhelming advantage. Even when face-to-face with an enemy, surprise can still be employed by attacking where he least expects it. To do this you must create an expectation in the enemy's mind through the use of a feint. If you plan to attack on the right flank, you first maneuver your left. If you wish to invade, you first pretend to improve your defense, if you intend to hold your ground, make a display of packing up.
\item Connection to HLSMAC scenario: Use threatening allied units to feign an attack on the enemy main base, drawing their forces away for our real assault.
\end{itemize}

\item \textbf{Stratagem: Lure Your Enemy onto the Roof, then Take Away the Ladder (swct)}
\begin{itemize}
\item Pinyin: Shang wu chou ti
\item Explanation: When facing a powerful enemy's pursuit, avoid direct engagement, but instead design an elaborate trap to lure the enemies deeper, subsequently blocking their line of advance and destroying the entire force in a narrow valley.
\item Connection to HLSMAC scenario: Transport Sentries to high ground via Warp Prism, then use Force Fields to trap and destroy enemies.
\end{itemize}

\item \textbf{Stratagem: Replace the Beams with Rotten Timbers (tlhz)}
\begin{itemize}
\item Pinyin: Tou liang huan zhu
\item Explanation: Disrupt the enemy's formations, interfere with their methods of operations, change the rules in which they are used to following, and go against their standard training. In this way you remove the supporting pillar, the common link which makes a group of men an effective fighting force.
\item Connection to HLSMAC scenario: Replace the expected Hatchery with an Evolution Chamber to disrupt enemy expectations and gain advantage through unexpected allied unit spawns.
\end{itemize}
\item \textbf{Stratagem: Besiege Wei to Rescue Zhao (wwjz)}
\begin{itemize}
\item Pinyin: Wei wei jiu zhao
\item Explanation: When the enemy is too strong to attack directly, strike at something he holds dear.
\item Connection to HLSMAC scenario: Attack the enemy base to force them to abandon their siege, then strike the scattered returning forces.
\end{itemize}
\item \textbf{Stratagem: Create Something from Nothing (wzsy)}
\begin{itemize}
\item Pinyin: Wu zhong sheng you
\item Explanation: Use the feint that the enemy will be hesitant to react to, and catch your enemy with his guard down.
\item Connection to HLSMAC scenario: Use Sentry's Hallucination ability to create illusory allies, forcing enemy retreat and creating opportunities to strike undefended targets.
\end{itemize}
\item \textbf{Stratagem: To Catch Something, First Let It Go (yqgz)}
\begin{itemize}
\item Pinyin: Yu qin gu zong
\item Explanation: Cornered prey will often mount a final desperate attack. To prevent this you let the enemy believe he still has a chance for freedom. His will to fight is thus dampened by his desire to escape. When in the end the freedom is proven a falsehood, the enemy's morale will be defeated and he will surrender without a fight.
\item Connection to HLSMAC scenario: Bait enemy advance with sacrificial units, then attack when their defenses are weakened.
\end{itemize}
\end{enumerate}

\subsection{HLSMAC Scenarios}
\label{sec:scenarios_appendix}

\begin{figure*}[htbp]
\centering
\includegraphics[width=2\columnwidth]{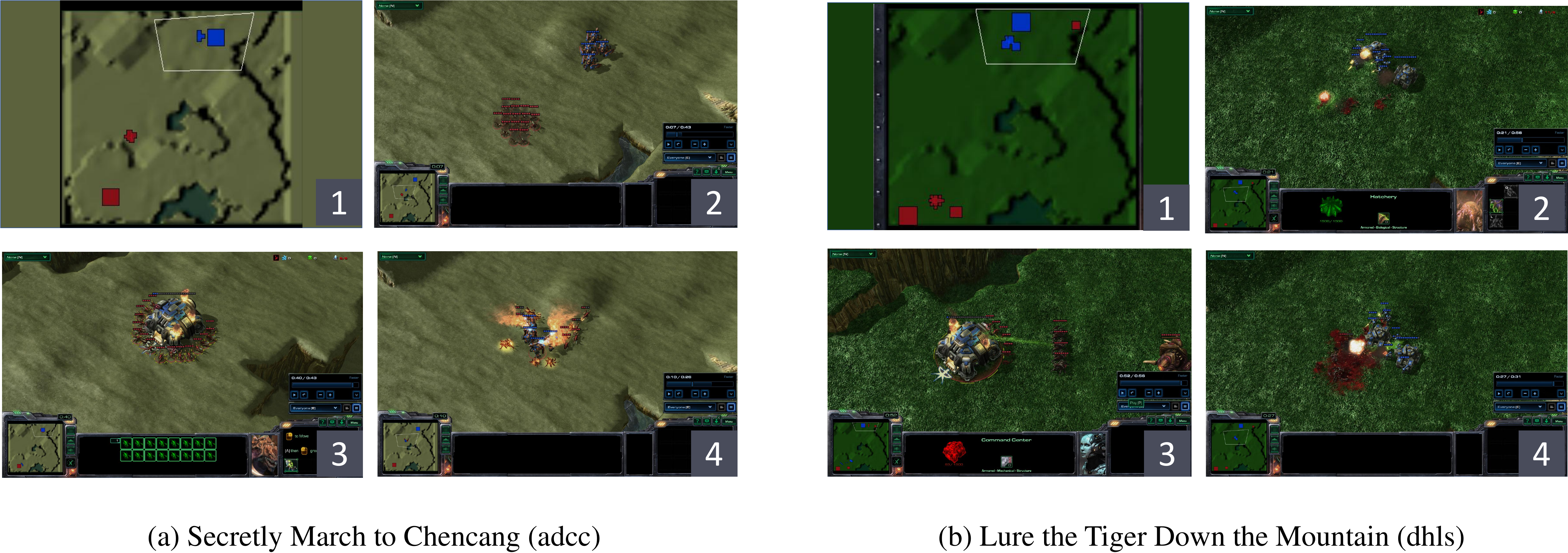} 
\caption{\textbf{adcc} and \textbf{dhls} scenarios. Scene 1 of each subfigure shows the initial battlefield (red: our forces, blue: enemy). Scenes 2-4 depict the early stage, the victory stage (with stratagem), and the defeat stage (without stratagem via direct combat).}
\label{fig:adcc_dhls}
\end{figure*}

\begin{figure*}[htbp]
\centering
\includegraphics[width=2\columnwidth]{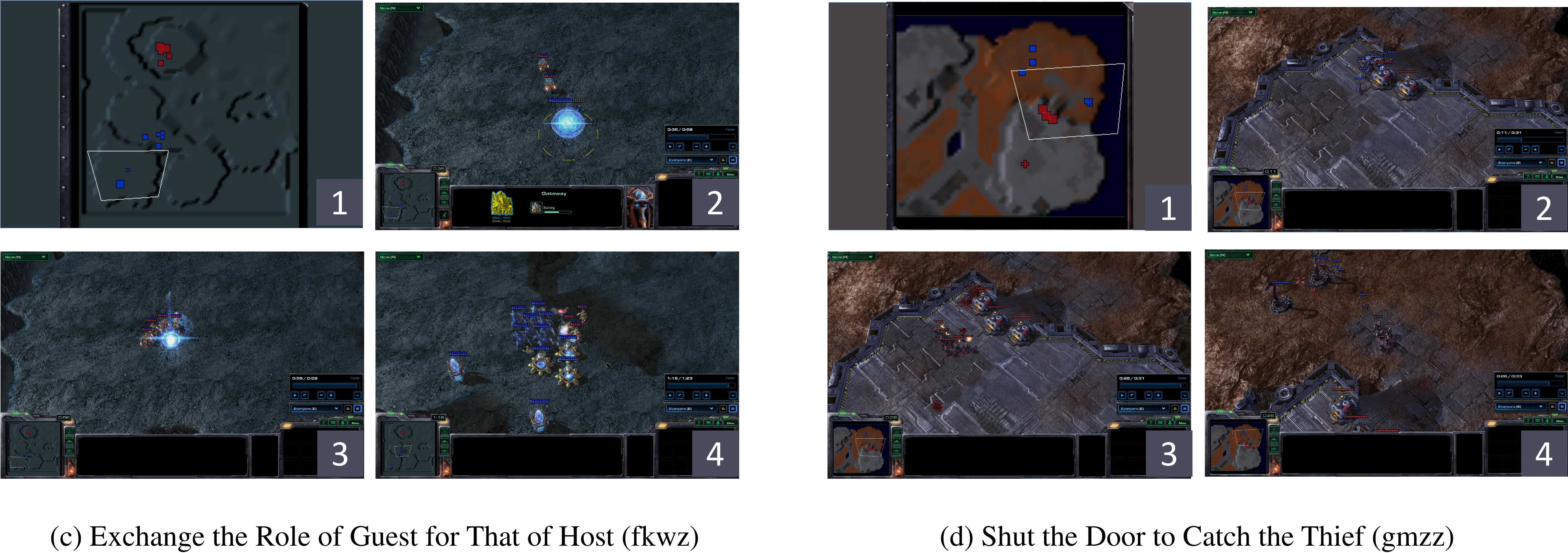} 
\caption{\textbf{fkwz} and \textbf{gmzz} scenarios. Scene 1 of each subfigure shows the initial battlefield (red: our forces, blue: enemy). Scenes 2-4 depict the early stage, the victory stage (with stratagem), and the defeat stage (without stratagem via direct combat).}
\label{fig:fkwz_gmzz}
\end{figure*}

\begin{figure*}[htbp]
\centering
\includegraphics[width=2\columnwidth]{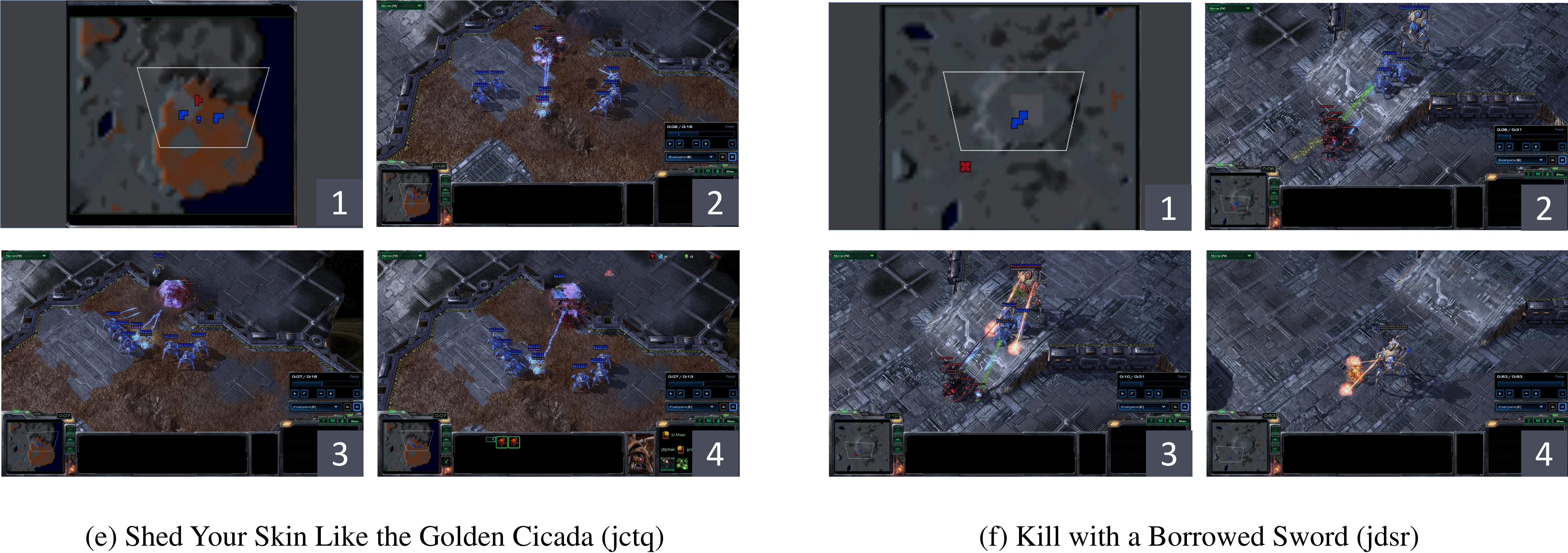} 
\caption{\textbf{jctq} and \textbf{jdsr} scenarios. Scene 1 of each subfigure shows the initial battlefield (red: our forces, blue: enemy). Scenes 2-4 depict the early stage, the victory stage (with stratagem), and the defeat stage (without stratagem via direct combat).}
\label{fig:jctq_jdsr}
\end{figure*}

\begin{figure*}[htbp]
\centering
\includegraphics[width=2\columnwidth]{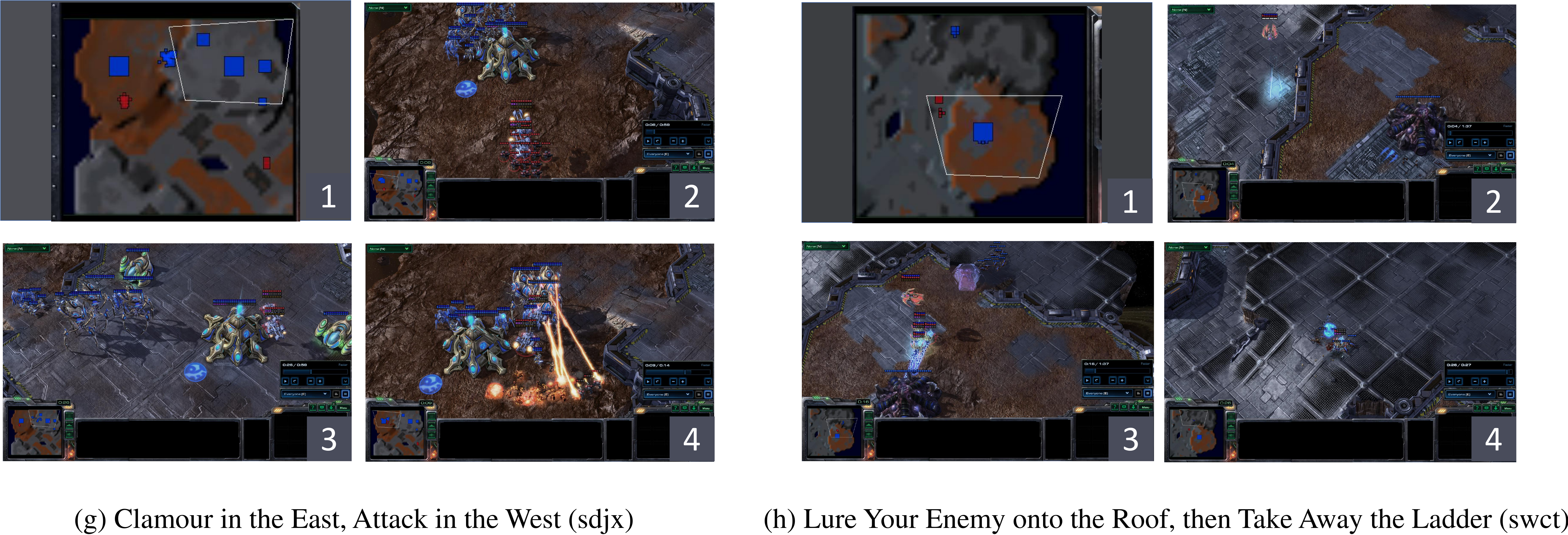} 
\caption{\textbf{sdjx} and \textbf{swct} scenarios. Scene 1 of each subfigure shows the initial battlefield (red: our forces, blue: enemy). Scenes 2-4 depict the early stage, the victory stage (with stratagem), and the defeat stage (without stratagem via direct combat).}
\label{fig:sdjx_swct}
\end{figure*}

\begin{figure*}[htbp]
\centering
\includegraphics[width=2\columnwidth]{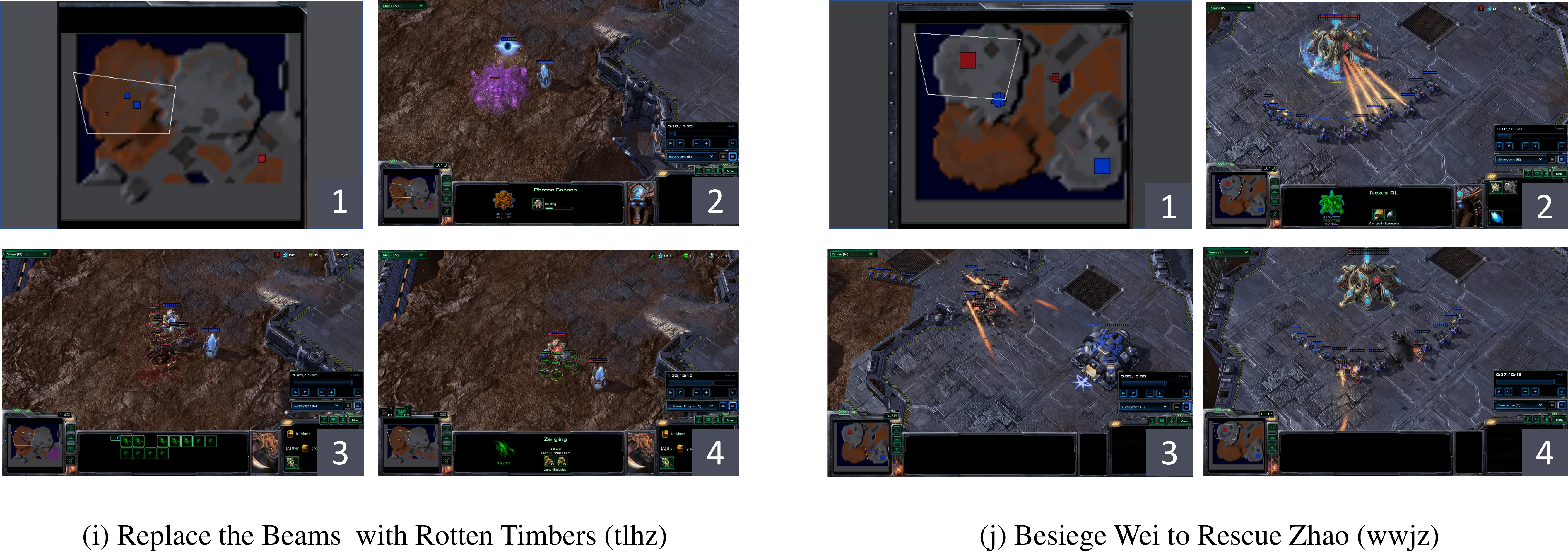} 
\caption{\textbf{tlhz} and \textbf{wwjz} scenarios. Scene 1 of each subfigure shows the initial battlefield (red: our forces, blue: enemy). Scenes 2-4 depict the early stage, the victory stage (with stratagem), and the defeat stage (without stratagem via direct combat).}
\label{fig:tlhz_wwjz}
\end{figure*}

\begin{figure*}[htbp]
\centering
\includegraphics[width=2\columnwidth]{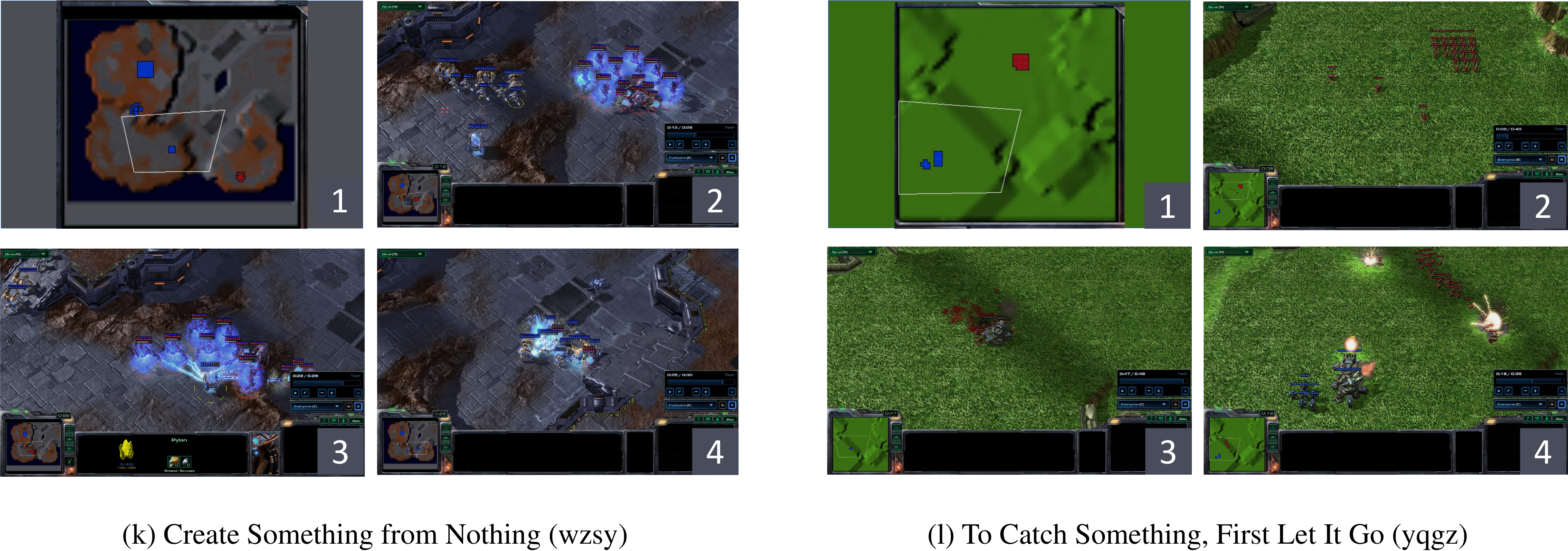} 
\caption{\textbf{wzsy} and \textbf{yqgz} scenarios. Scene 1 of each subfigure shows the initial battlefield (red: our forces, blue: enemy). Scenes 2-4 depict the early stage, the victory stage (with stratagem), and the defeat stage (without stratagem via direct combat).}
\label{fig:wzsy_yqgz}
\end{figure*}

This section provides comprehensive descriptions of the 12 HLSMAC scenarios. We detail the unit compositions, special abilities, game mechanisms, termination conditions, and expected solutions following the stratagem. These well-designed scenarios aim to evaluate multi-agent coordination and high-level strategic decision-making. For a more intuitive understanding of each scenario, please refer to Figures~\ref{fig:adcc_dhls}-~\ref{fig:wzsy_yqgz}.

\begin{enumerate}
\item \textbf{Scenario: adcc (Secretly March to Chencang)}
\begin{itemize}
    \item Units:
    \begin{itemize}
        \item[--] Ally: 16 Zerglings, 1 Hatchery.
        \item[--] Enemy: 4 Hellbats, 1 Command Center.
    \end{itemize}
    \item Special abilities:
    \begin{itemize}
        \item[--] \texttt{Burrow}/\texttt{Unburrow} for allied Zerglings.
    \end{itemize}
    \item Game mechanisms:
    \begin{itemize}
        \item[--] When the game starts, the enemy Hellbats will advance from the Command Center to attack our Hatchery.
        \item[--] If we engage the enemy directly, we will likely lose.
    \end{itemize}
    \item Termination conditions:
    \begin{itemize}
        \item[--] Win if the enemy's Command Center is destroyed, or all enemy units are killed.
        \item[--] Lose if our Hatchery is destroyed, or all our Zerglings are killed.
    \end{itemize}
    \item Expected solutions:
    \begin{itemize}
        \item[--] Our Zerglings should either use the Burrow ability to avoid damage or take an alternative route to bypass the enemy Hellbats.
        \item[--] Our Zerglings advance to attack the enemy Command Center after successfully avoiding the Hellbats.
    \end{itemize}
\end{itemize}

\item \textbf{Scenario: dhls (Lure the Tiger Down the Mountain)}
\begin{itemize}
    \item Units:
    \begin{itemize}
        \item[--] Ally: 9 Zerglings, 4 Roaches, 1 Hatchery, 1 Nydus Network, 1 Nydus Worm.
        \item[--] Enemy: 8 Marines, 2 Siege Tanks, 1 Command Center.
    \end{itemize}
    \item Special abilities:
    \begin{itemize}
        \item[--] \texttt{Load} for allied Nydus Network.
        \item[--] \texttt{Unload} for allied Nydus Worm.
    \end{itemize}
    \item Game mechanisms:
    \begin{itemize}
        \item[--] The enemy forces are defending their base.
        \item[--] If any of our units dies during the engagement, the enemy forces will advance from the Command Center to attack our Hatchery.
        \item[--] If we engage the enemy directly, we will likely lose.
    \end{itemize}
    \item Termination conditions:
    \begin{itemize}
        \item[--] Win if the enemy's Command Center is destroyed, or all enemy units are killed.
        \item[--] Lose if our Hatchery is destroyed, or all our units are killed.
    \end{itemize}
    \item Expected solutions:
    \begin{itemize}
        \item[--] A small group of our units lures the enemy forces away.
        \item[--] Our main force attacks the enemy base.
    \end{itemize}
\end{itemize}

\item \textbf{Scenario: fkwz (Exchange the Role of Guest for That of Host)}
\begin{itemize}
    \item Units:
    \begin{itemize}
        \item[--] Ally: 2 Warp Gates, 2 Pylons, 1 Warp Prism.
        \item[--] Enemy: 1 Stalker, 1 Gateway (in building progress), 2 Pylons, 3 Photon Cannons.
    \end{itemize}
    \item Special abilities:
    \begin{itemize}
        \item[--] \texttt{Warp in Zealot} for allied Warp Gates.
        \item[--] \texttt{Load/Unload} and \texttt{Phasing/Transport} for allied Warp Prism.
    \end{itemize}
    \item Game mechanisms:
    \begin{itemize}
        \item[--] When the game starts, the enemy Stalker will advance from the Gateway to attack our Warp Gates.
        \item[--] If the enemy completes the Gateway construction, six additional enemy Stalkers will spawn and advance from the Gateway to attack our Warp Gates.
        \item[--] If we attack via the main path blocked by the enemy Photon Cannons, we will likely lose.
    \end{itemize}
    \item Termination conditions:
    \begin{itemize}
        \item[--] Win if all enemy Stalkers are killed, or enemy's Gateway is destroyed.
        \item[--] Lose if any of our Warp Gates are destroyed, or our Warp Prism is killed.
    \end{itemize}
    \item Expected solutions:
    \begin{itemize}
        \item[--] Our Warp Gates warp in two Zealots at our base.
        \item[--] Our Warp Prism transports the two Zealots to the enemy base.
        \item[--] Our Warp Prism enters Phasing mode to warp in two additional Zealots near the enemy base.
        \item[--] All Zealots destroy the enemy Gateway.
    \end{itemize}
\end{itemize}

\item \textbf{Scenario: gmzz (Shut the Door to Catch the Thief)}
\begin{itemize}
    \item Units:
    \begin{itemize}
        \item[--] Ally: 5 Marines, 3 Supply Depots.
        \item[--] Enemy: 8 Zerglings, 3 Spine Crawlers.
    \end{itemize}
    \item Special abilities:
    \begin{itemize}
        \item[--] \texttt{SupplyDepotLower}/\texttt{SupplyDepotRaise} for allied Supply Depots.
    \end{itemize}
    \item Game mechanisms:
    \begin{itemize}
        \item[--] When the game starts, the enemy Zerglings are positioned outside, walled off by our Supply Deopts.
        \item[--] If we lower any Supply Depot, the enemy Zerglings will attempt to reach the high ground and attack our Marines.
        \item[--] If two or more enemy Zerglings die, the remaining Zerglings will retreat to the Spine Crawlers.
        \item[--] Without the support of the enemy Spine Crawlers, our Marines can defeat the enemy Zerglings. With their support, we will likely lose.
    \end{itemize}
    \item Termination conditions:
    \begin{itemize}        
        \item[--] Win if all enemy Zerglings are killed.
        \item[--] Lose if all our Marines are killed.
    \end{itemize}
    \item Expected solutions:
    \begin{itemize}
        \item[--] Lower Supply Depots to lure the enemy Zerglings onto our high ground.
        \item[--] Raise Supply Depots before the enemy Zerglings retreat.
        \item[--] Our Marines eliminate the trapped Zerglings.
    \end{itemize}
\end{itemize}

\item \textbf{Scenario: jctq (Shed Your Skin Like the Golden Cicada)}
\begin{itemize}
    \item Units:
    \begin{itemize}
        \item[--] Ally: 4 Roaches.
        \item[--] Enemy: 2 Sentries, 6 Stalkers, 1 Observer.
    \end{itemize}
    \item Special abilities:
    \begin{itemize}
        \item[--] \texttt{Burrow}/\texttt{Unburrow} for allied Roaches.
    \end{itemize}
    \item Game mechanisms:
    \begin{itemize}
        \item[--] When the game starts, the enemy forces will attack our Roaches trapped on the high ground with only one exit.
        \item[--] If any of our Roaches attempts to escape, the enemy Sentries will cast the Force Field ability to block the exit.
        \item[--] We will likely lose all our forces.
    \end{itemize}
    \item Termination conditions:
    \begin{itemize}
        \item[--] Win if any of our roaches remain alive after timeout, or all enemy units are killed.
        \item[--] Lose if all our roaches are killed.
    \end{itemize}
    \item Expected solutions:
    \begin{itemize}
        \item[--] Our Roaches burrow while under attack. 
        \item[--] Our Roaches move through the ramp while burrowed to escape encirclement as quickly as possible.
    \end{itemize}
\end{itemize}

\item \textbf{Scenario: jdsr (Kill with a Borrowed Sword)}
\begin{itemize}
    \item Units:
    \begin{itemize}
        \item[--] Ally: 4 Roaches, 1 Infestor.
        \item[--] Enemy: 3 Stalkers, 1 Colossus.
    \end{itemize}
    \item Special abilities:
    \begin{itemize}
        \item[--] \texttt{Neural Parasite} for allied Infestor.
    \end{itemize}
    \item Game mechanisms:
    \begin{itemize}
        \item[--] When the game starts, the enemy forces will advance to attack us proactively.
        \item[--] Without utilizing unit abilities, we will likely lose.
    \end{itemize}
    \item Termination conditions:
    \begin{itemize}
        \item[--] Win if all enemy units are killed.
        \item[--] Lose if all our units are killed.
    \end{itemize}
    \item Expected solutions:
    \begin{itemize}
        \item[--] Our Infestor casts Neural Parasite to control the enemy Colossus, the most powerful enemy unit.
        \item[--] All our forces attack the enemy forces.
    \end{itemize}
\end{itemize}

\item \textbf{Scenario: sdjx (Clamour in the East, Attack in the West)}
\begin{itemize}
    \item Units:
    \begin{itemize}
        \item[--] Ally: 14 Marines, 4 Medivacs.
        \item[--] Enemy: 5 Zealots, 4 Stalkers, 3 Colossi, 2 Nexuses, 2 Assimilators, 1 Pylon.
    \end{itemize}
    \item Special abilities:
    \begin{itemize}
        \item[--] \texttt{Healing} for allied Medivacs.
    \end{itemize}
    \item Game mechanisms:
    \begin{itemize}
        \item[--] When the game starts, the enemy forces are heavily guarding their expansion base while leaving the main base, located on the high ground, undefended.
        \item[--] If our units approach the enemy's main base, the enemy forces will retreat to defend it, leaving the expansion base undefended.
        \item[--] If we engage the enemy directly, we will likely lose.
    \end{itemize}
    \item Termination conditions:
    \begin{itemize}
        \item[--] Win if any of the enemy's Nexuses is destroyed, or all enemy units are killed.
        \item[--] Lose if all our Marines are killed.
    \end{itemize}
    \item Expected solutions:
    \begin{itemize}      
        \item[--] Two Medivacs feign an attack on the enemy's main base, causing the enemy forces to leave their expansion base.
        \item[--] Our main force attacks the enemy's expansion base.
    \end{itemize}
\end{itemize}

\item \textbf{Scenario: swct (Lure Your Enemy onto the Roof, then Take Away the Ladder)}
\begin{itemize}
    \item Units:
    \begin{itemize}
        \item[--] Ally: 4 Sentries, 1 Warp Prism.
        \item[--] Enemy: 10 Zerglings, 1 Hatchery.
    \end{itemize}
    \item Special abilities:
    \begin{itemize}
        \item[--] \texttt{Force Field} for allied Sentries.
        \item[--] \texttt{Load}/\texttt{Unload} for allied Warp Prism.
    \end{itemize}
    \item Game mechanisms:
    \begin{itemize}
        \item[--] If our units approach the enemy Hatchery, the enemy Zerglings will retreat to defend it.
        \item[--] If we engage the enemy directly, we will likely lose.
    \end{itemize}
    \item Termination conditions:
    \begin{itemize}
        \item[--] Win if the enemy's Hatchery is destroyed, or all enemy Zerglings are killed.
        \item[--] Lose if all our Sentries are killed.
    \end{itemize}
    \item Expected solutions:
    \begin{itemize}
        \item[--] Our Warp Prism loads all Sentries and drops them on the enemy's high ground.
        \item[--] Our Sentries maintain Force Fields on the ramp to block enemy advances.
        \item[--] Our Sentries attack the enemy.
    \end{itemize}
\end{itemize}

\item \textbf{Scenario: tlhz (Replace the Beams with Rotten Timbers)}
\begin{itemize}
    \item Units:
    \begin{itemize}
        \item[--] Ally: 1 Drone, 3 Larvas.
        \item[--] Enemy: 1 Photon Cannon (in building progress), 1 Pylon.
    \end{itemize}
    \item Special abilities:
    \begin{itemize}
        \item[--] \texttt{Mutate into Hatchery}, \texttt{Mutate into Evolution Chamber}, \texttt{CancelBuilding} for allied Drone.
        \item[--] \texttt{Morph to Zergling} for allied Larvas.
    \end{itemize}
    \item Game mechanisms:
    \begin{itemize}
        \item[--] A Hatchery provides creep in the nearby area.
        \item[--] We can only build an Evolution Chamber on creep.
        \item[--] When our Evolution Chamber is destroyed, six additional Broodlings will spawn, reinforcing our army.
    \end{itemize}
    \item Termination conditions:
    \begin{itemize}
        \item[--] Win if the enemy's Photon Canon is destroyed, or the enemy's Pylon is destroyed.
        \item[--] Lose if all our units and structures are killed.
    \end{itemize}
    \item Expected solutions:
    \begin{itemize}
        \item[--] Our Drone builds a Hatchery near the Photon Cannon.
        \item[--] Cancel the in-progress Hatchery, then build an Evolution Chamber on the creep that the Hatchery generated.
        \item[--] Our three Larvas morph into six Zerglings.
        \item[--] After the enemy destroys our Evolution Chamber, all our forces attack the enemy Photon Cannon and Pylon.
    \end{itemize}
\end{itemize}

\item \textbf{Scenario: wwjz (Besiege Wei to Rescue Zhao)}
\begin{itemize}
    \item Units:
    \begin{itemize} 
        \item[--] Ally: 7 Zealots, 1 Nexus.        
        \item[--] Enemy: 8 Hellions, 6 Marines, 1 Command Center.
    \end{itemize}
    \item Game mechanisms:
    \begin{itemize}
        \item[--] When the game starts, the enemy forces are attacking our Nexus.
        \item[--] Our Zealots cannot reach our Nexus in time to defend it. Even if we successfully retreat and engage the enemy directly, we will likely lose.
        \item[--] If our Zealots approach the enemy Command Center, the enemy forces will retreat to defend the Command Center.
    \end{itemize}
    \item Termination conditions:
    \begin{itemize}
        \item[--] Win if the enemy's Command Center is destroyed, or all enemy units are killed.
        \item[--] Lose if our Nexus is destroyed, or all our Zealots are killed.
    \end{itemize}
    \item Expected solutions:
    \begin{itemize}
        \item[--] Our Zealots encircle the enemy base, forcing the enemy forces to retreat and defend it.
        \item[--] Our Zealots attack the returning, scattered enemy forces.
    \end{itemize}
\end{itemize}

\item \textbf{Scenario: wzsy (Create Something from Nothing)}
\begin{itemize}
    \item Units:
    \begin{itemize} 
        \item[--] Ally: 2 Stalkers, 4 Sentries.
        \item[--] Enemy: 3 Immortals, 5 Zealots, 1 Nexus, 1 Pylon.
    \end{itemize}
    \item Special abilities:
    \begin{itemize}
        \item[--] \texttt{Hallucination} for allied Sentries.
    \end{itemize}
    \item Game mechanisms:
    \begin{itemize}
        \item[--] When the game starts, the enemy forces will advance from the Nexus to attack us proactively.
        \item[--] When ten or more of our units enter enemy sight range, the enemy forces will retreat to the Nexus.
        \item[--] If we engage the enemy directly, we will likely lose.
    \end{itemize}
    \item Termination conditions:
    \begin{itemize}
        \item[--] Win if the enemy's pylon is destroyed, or all enemy units are killed, or the enemy's Nexus is destroyed.
        \item[--] Lose if all our units are killed.
    \end{itemize}
    \item Expected solutions:
    \begin{itemize}
        \item[--] Our Sentries cast Hallucination to scare off enemy forces.
        \item[--] All our forces attack the enemy Pylon on the midway, and even further, defeat the retreating enemy forces.
    \end{itemize}
\end{itemize}

\item \textbf{Scenario: yqgz (To Catch Something, First Let It Go)}
\begin{itemize}
    \item Units:
    \begin{itemize} 
        \item[--] Ally: 24 Zerglings.
        \item[--] Enemy: 6 Marines, 2 Siege Tanks (Siege Mode).
    \end{itemize}
    \item Game mechanisms:
    \begin{itemize}
        \item[--] When the game starts, the enemy forces are in defensive formation. If we engage the enemy directly at this time, we will likely lose.
        \item[--] If the enemy Siege Tanks kill at least three of our Zerglings within their sight range, they will switch to Tank Mode. Then the enemy forces advance to attack us.
    \end{itemize}
    \item Termination conditions:
    \begin{itemize}
        \item[--] Win if all enemy units are killed.
        \item[--] Lose if all our Zerglings are killed.
    \end{itemize}
    \item Expected solutions:
    \begin{itemize}
        \item[--] A small group of our units lures the enemy into a proactive attack.
        \item[--] Our main force attacks the enemy forces.
    \end{itemize}
\end{itemize}
\end{enumerate}

\subsection{Experimental Setup}
\subsubsection{About MARL baselines}
We conduct a comprehensive evaluation of 21 state-of-the-art multi-agent reinforcement learning algorithms. To facilitate a fair comparison, we standardized common hyperparameters while preserving each algorithm's unique characteristics through its specific parameter configurations. All experiments are conducted on StarCraft II with 12 HLSMAC scenarios at difficulty 7. The shared standard HLSMAC environment configurations are detailed in Table ~\ref{table:hlsmac_parameters}. 
Table ~\ref{table:common_parameters} presents the common parameters and their values shared across all algorithms, while Tables ~\ref{table:iql_parameters} - \ref{table:dtape_parameters} detail the specific parameter configurations for each individual algorithm.

Agents receive rich observational information, including their own health points, ally health information, terrain height, and pathing grids. Notably, agents do not observe their previous actions or timestep numbers, requiring them to maintain internal memory through recurrent networks.

The environment uses a balanced reward function with victory rewards of 200 points and death penalties of 10 points. Reward scaling is enabled with a rate of 20 and negative scaling of 0.5. The reward system requires agents to balance offensive gains against potential losses by setting the `reward\_only\_positive' to False.

The environment runs with a step multiplier of 8 to balance simulation speed and control granularity, while agents can move up to 2 units per action. All heuristic AI assistance is disabled to ensure pure multi-agent learning.

Training proceeds for 2,000,000 timesteps with evaluation every 10,000 steps using 32 episodes in decentralized execution mode.

\begin{table}[htbp]
	\centering
	\begin{tabular}{|p{0.5\columnwidth}|p{0.4\columnwidth}|}
	\hline
	Parameter & Value \\ 
	\hline
	continuing\_episode & False \\
	debug & False \\
	difficulty & 7 \\
	game\_version & None \\
	heuristic\_ai & False \\
	heuristic\_rest & False \\
	map\_name & HLSMAC's 12 maps \\
	move\_amount & 2 \\
	obs\_all\_health & True \\
	obs\_instead\_of\_state & False \\
	obs\_last\_action & False \\
	obs\_own\_health & True \\
	obs\_pathing\_grid & True \\
	obs\_terrain\_height & True \\
	obs\_timestep\_number & False \\
	reward\_death\_value & 10 \\
	reward\_defeat & 0 \\
	reward\_negative\_scale & 0.5 \\
	reward\_only\_positive & False \\
	reward\_scale & True \\
	reward\_scale\_rate & 20 \\
	reward\_sparse & False \\
	reward\_win & 200 \\
	seed & e.g., 100845290 \\
	state\_last\_action & True, False \\
	state\_timestep\_number & False \\
	step\_mul & 8 \\
	\hline
	\end{tabular}
	\caption{HLSMAC environment parameter configurations}
	\label{table:hlsmac_parameters}
\end{table}

\begin{table}[htbp]
\centering
\begin{tabular}{|p{0.5\columnwidth}|p{0.4\columnwidth}|}
\hline
Parameter & Value \\
\hline
rnn\_hidden\_dim & 64 \\
mixing\_embed\_dim & 32,64 \\
agent\_output\_type & pi\_logits, q \\
action\_selector & epsilon\_greedy, gumbel, \\
& multinomial, soft\_epsilon\_greedy \\
epsilon\_start & 0--1 \\
epsilon\_finish & 0, 0.01, 0.05 \\
epsilon\_anneal\_time & 50--500000 \\
test\_greedy & True \\
lr & 0.00025--0.0025 \\
critic\_lr & 0.0001, 0.0005, 0.001 \\
gamma & 0.99 \\
grad\_norm\_clip & 10, 20 \\
target\_update\_interval & 200, 600 \\
optim\_alpha & 0.99 \\
optim\_eps & 1e-5 \\
batch\_size & 8-128 \\
buffer\_size & 8-5000 \\
buffer\_cpu\_only & False, True \\
runner & episode, parallel \\
batch\_size\_run & 1 \\
t\_max & 2005000 \\
use\_cuda & True \\
obs\_agent\_id & True \\
obs\_last\_action & True \\
evaluate & True \\
test\_interval & 10000 \\
test\_nepisode & 32 \\
log\_interval & 10000 \\
learner\_log\_interval & 10000 \\
runner\_log\_interval & 10000 \\
save\_model & True \\
save\_model\_interval & 2000000 \\
save\_replay & False \\
local\_results\_path & results \\
label & default\_label \\
repeat\_id & 1 \\
load\_step & 0 \\
\hline
\end{tabular}
\caption{Common parameters across all algorithms}
\label{table:common_parameters}
\end{table}

\begin{table}[htbp]
\centering
\begin{tabular}{|p{0.5\columnwidth}|p{0.4\columnwidth}|}
\hline
Parameter & Value \\
\hline
agent & rnn \\
double\_q & True \\
learner & q\_learner \\
mac & basic\_mac \\
mixer & None \\
\hline
\end{tabular}
\caption{IQL algorithm specific parameters}
\label{table:iql_parameters}
\end{table}

\begin{table}[htbp]
\centering
\begin{tabular}{|p{0.5\columnwidth}|p{0.4\columnwidth}|}
\hline
Parameter & Value \\
\hline
agent & rnn \\
critic\_baseline\_fn & coma \\
critic\_q\_fn & coma \\
critic\_train\_mode & seq \\
critic\_train\_reps & 1 \\
learner & coma\_learner \\
mac & basic\_mac \\
mask\_before\_softmax & False \\
q\_nstep & 0 \\
td\_lambda & 0.8 \\
\hline
\end{tabular}
\caption{COMA algorithm specific parameters}
\label{table:coma_parameters}
\end{table}

\begin{table}[htbp]
\centering
\begin{tabular}{|p{0.5\columnwidth}|p{0.4\columnwidth}|}
\hline
Parameter & Value \\
\hline
agent & rnn \\
double\_q & True \\
learner & q\_learner \\
mac & basic\_mac \\
mixer & vdn \\
\hline
\end{tabular}
\caption{VDN algorithm specific parameters}
\label{table:vdn_parameters}
\end{table}

\begin{table}[htbp]
\centering
\begin{tabular}{|p{0.5\columnwidth}|p{0.4\columnwidth}|}
\hline
Parameter & Value \\
\hline
agent & rnn \\
double\_q & True \\
hypernet\_embed & 64 \\
hypernet\_layers & 2 \\
learner & q\_learner \\
mac & basic\_mac \\
mixer & qmix \\
\hline
\end{tabular}
\caption{QMIX algorithm specific parameters}
\label{table:qmix_parameters}
\end{table}

\begin{table}[htbp]
\centering
\begin{tabular}{|p{0.5\columnwidth}|p{0.4\columnwidth}|}
\hline
Parameter & Value \\
\hline
abs & True \\
agent & rnn\_ppo \\
ent\_coef & 0.01 \\
hypernet\_embed & 64 \\
hypernet\_layers & 2 \\
lam & 0.95 \\
learner & policy\_gradient \\
mac & ppo\_mac \\
mask\_before\_softmax & True \\
mixer & qmix \\
optim & rmsprop \\
q\_nstep & 0 \\
run & default \\
td\_lambda & 0.8 \\
vf\_coef & 0.1 \\
\hline
\end{tabular}
\caption{VMIX algorithm specific parameters}
\label{table:vmix_parameters}
\end{table}

\begin{table}[htbp]
\centering
\begin{tabular}{|p{0.5\columnwidth}|p{0.4\columnwidth}|}
\hline
Parameter & Value \\
\hline
agent & noise\_rnn \\
bandit\_batch & 64 \\
bandit\_buffer & 512 \\
bandit\_epsilon & 0.1 \\
bandit\_iters & 8 \\
bandit\_policy & True \\
bandit\_reward\_scaling & 20 \\
bandit\_use\_state & True \\
discrim\_layers & 1 \\
discrim\_size & 32 \\
double\_q & True \\
entropy\_scaling & 0.001 \\
hard\_qs & False \\
hyper\_initialization\_nonzeros & 0 \\
learner & noise\_q\_learner \\
mac & noise\_mac \\
mi\_intrinsic & False \\
mi\_loss & 1 \\
mi\_scaler & 0.1 \\
mixer & qmix \\
noise\_bandit & False \\
noise\_dim & 2 \\
noise\_embedding\_dim & 32 \\
recurrent\_critic & False \\
rnn\_agg\_size & 32 \\
rnn\_discrim & False \\
skip\_connections & False \\
\hline
\end{tabular}
\caption{MAVEN algorithm specific parameters}
\label{table:maven_parameters}
\end{table}

\begin{table}[htbp]
\centering
\begin{tabular}{|p{0.5\columnwidth}|p{0.4\columnwidth}|}
\hline
Parameter & Value \\
\hline
agent & rnn \\
double\_q & True \\
learner & qtran\_learner \\
mac & basic\_mac \\
mixer & qtran\_base \\
network\_size & small \\
nopt\_min\_loss & 0.1 \\
opt\_loss & 1 \\
qtran\_arch & qtran\_paper \\
\hline
\end{tabular}
\caption{QTRAN algorithm specific parameters}
\label{table:qtran_parameters}
\end{table}

\begin{table}[htbp]
\centering
\begin{tabular}{|p{0.5\columnwidth}|p{0.4\columnwidth}|}
\hline
Parameter & Value \\
\hline
agent & rnn \\
central\_action\_embed & 1 \\
central\_agent & central\_rnn \\
central\_loss & 1 \\
central\_mac & basic\_central\_mac \\
central\_mixer & ff \\
central\_mixing\_embed\_dim & 256 \\
central\_rnn\_hidden\_dim & 64 \\
double\_q & True \\
gated & False \\
hypernet\_embed & 64 \\
hypernet\_layers & 2 \\
hysteretic\_qmix & False \\
learner & max\_q\_learner \\
mac & basic\_mac \\
mixer & qmix \\
qmix\_loss & 1 \\
recurrent\_critic & False \\
training\_iters & 1 \\
w & 0.1 \\
\hline
\end{tabular}
\caption{CWQMIX algorithm specific parameters}
\label{table:cwqmix_parameters}
\end{table}

\begin{table}[htbp]
\centering
\begin{tabular}{|p{0.5\columnwidth}|p{0.4\columnwidth}|}
\hline
Parameter & Value \\
\hline
agent & rnn \\
central\_action\_embed & 1 \\
central\_agent & central\_rnn \\
central\_loss & 1 \\
central\_mac & basic\_central\_mac \\
central\_mixer & ff \\
central\_mixing\_embed\_dim & 256 \\
central\_rnn\_hidden\_dim & 64 \\
double\_q & True \\
gated & False \\
hypernet\_embed & 64 \\
hypernet\_layers & 2 \\
hysteretic\_qmix & True \\
learner & max\_q\_learner \\
mac & basic\_mac \\
mixer & qmix \\
qmix\_loss & 1 \\
recurrent\_critic & False \\
training\_iters & 1 \\
w & 0.1 \\
\hline
\end{tabular}
\caption{OWQMIX algorithm specific parameters}
\label{table:owqmix_parameters}
\end{table}

\begin{table}[htbp]
\centering
\begin{tabular}{|p{0.5\columnwidth}|p{0.4\columnwidth}|}
\hline
Parameter & Value \\
\hline
agent & rnn \\
critic\_baseline\_fn & coma \\
critic\_q\_fn & coma \\
critic\_train\_mode & seq \\
critic\_train\_reps & 1 \\
learner & offpg\_learner \\
mac & basic\_mac \\
mask\_before\_softmax & False \\
off\_batch\_size & 32 \\
off\_buffer\_size & 2000 \\
q\_nstep & 0 \\
step & 5 \\
tb\_lambda & 0.93 \\
td\_lambda & 0.8 \\
\hline
\end{tabular}
\caption{DOP algorithm specific parameters}
\label{table:dop_parameters}
\end{table}

\begin{table}[htbp]
\centering
\begin{tabular}{|p{0.5\columnwidth}|p{0.4\columnwidth}|}
\hline
Parameter & Value \\
\hline
agent & rnn \\
critic & lica \\
critics\_update\_num & 1 \\
entropy\_coef & 0.06 \\
hypernet\_embed\_dim & 64 \\
hypernet\_layers & 2 \\
learner & lica\_learner \\
lica\_mixing\_embed\_dim & 64 \\
mac & lica\_mac \\
mask\_before\_softmax & True \\
run & default \\
td\_lambda & 0.6 \\
\hline
\end{tabular}
\caption{LICA algorithm specific parameters}
\label{table:lica_parameters}
\end{table}

\begin{table}[htbp]
\centering
\begin{tabular}{|p{0.5\columnwidth}|p{0.4\columnwidth}|}
\hline
Parameter & Value \\
\hline
adv\_hypernet\_embed & 64 \\
adv\_hypernet\_layers & 1 \\
agent & rnn \\
attend\_reg\_coef & 0.001 \\
burn\_in\_period & 100 \\
double\_q & True \\
gated & False \\
hypernet\_embed & 64 \\
is\_adv\_attention & True \\
is\_minus\_one & True \\
is\_stop\_gradient & True \\
learner & Qatten\_learner \\
mac & basic\_mac \\
mask\_dead & False \\
mixer & Qatten \\
n\_head & 4 \\
nonlinear & False \\
num\_kernel & 4 \\
recurrent\_critic & False \\
state\_bias & True \\
training\_iters & 1 \\
weighted\_head & False \\
\hline
\end{tabular}
\caption{Qatten algorithm specific parameters}
\label{table:Qatten_parameters}
\end{table}

\begin{table}[htbp]
\centering
\begin{tabular}{|p{0.5\columnwidth}|p{0.4\columnwidth}|}
\hline
Parameter & Value \\
\hline
adv\_hypernet\_embed & 64 \\
adv\_hypernet\_layers & 3 \\
agent & rnn \\
double\_q & True \\
gated & False \\
hypernet\_embed & 64 \\
is\_adv\_attention & True \\
is\_minus\_one & True \\
is\_stop\_gradient & True \\
learner & dmaq\_Qatten\_learner \\
mac & basic\_mac \\
mixer & dmaq \\
num\_kernel & 10 \\
recurrent\_critic & False \\
training\_iters & 1 \\
weighted\_head & True \\
\hline
\end{tabular}
\caption{QPLEX algorithm specific parameters}
\label{table:qplex_parameters}
\end{table}

\begin{table}[htbp]
\centering
\begin{tabular}{|p{0.5\columnwidth}|p{0.4\columnwidth}|}
\hline
Parameter & Value \\
\hline
agent & rnn \\
burn\_in\_period & 100 \\
c\_lr & 0.0005 \\
learner & fop\_learner \\
mac & basic\_mac \\
mask\_before\_softmax & False \\
n\_head & 4 \\
td\_lambda & 0.8 \\
\hline
\end{tabular}
\caption{FOP algorithm specific parameters}
\label{table:fop_parameters}
\end{table}

\begin{table}[htbp]
\centering
\begin{tabular}{|p{0.5\columnwidth}|p{0.4\columnwidth}|}
\hline
Parameter & Value \\
\hline
abs & True \\
agent & rnn \\
critic\_hidden\_dim & 128 \\
entropy\_coef & 0.03 \\
hypernet\_embed & 64 \\
lambd & 0.6 \\
learner & fmac\_learner \\
mac & lica\_mac \\
mask\_before\_softmax & True \\
mixer & qmix \\
name & riit\_env=8\_ada \\
off\_batch\_size & 64 \\
off\_buffer\_size & 5000 \\
optimizer & adam \\
run & on\_off \\
\hline
\end{tabular}
\caption{RIIT algorithm specific parameters}
\label{table:riit_parameters}
\end{table}

\begin{table}[htbp]
\centering
\begin{tabular}{|p{0.5\columnwidth}|p{0.4\columnwidth}|}
\hline
Parameter & Value \\
\hline
action\_encoder & obs\_reward \\
action\_latent\_dim & 20 \\
agent & rode \\
bi\_opt & False \\
double\_q & True \\
epsilon\_anneal\_time\_exp & 70000 \\
hypernet\_embed & 64 \\
hypernet\_layers & 2 \\
learner & rode\_learner \\
mac & rode\_mac \\
mixer & qmix \\
n\_role\_clusters & 5 \\
role & dot \\
role\_action\_spaces\_update\_start & 50000 \\
role\_epsilon\_finish & 0.05 \\
role\_interval & 5 \\
role\_mixer & qmix \\
role\_selector & dot \\
state\_latent\_dim & 32 \\
verbose & False \\
\hline
\end{tabular}
\caption{RODE algorithm specific parameters}
\label{table:rode_parameters}
\end{table}

\begin{table}[htbp]
\centering
\begin{tabular}{|p{0.5\columnwidth}|p{0.4\columnwidth}|}
\hline
Parameter & Value \\
\hline
agent & latent\_ce\_dis\_agent \\
device\_num & 0 \\
dis\_loss\_weight & 0.001 \\
dis\_sigmoid & False \\
dis\_time & 0 \\
double\_q & True \\
h\_loss\_weight & 0.0001 \\
kl\_loss\_weight & 0.0001 \\
latent\_dim & 3 \\
learner & latent\_q\_learner \\
mac & separate\_mac \\
mixer & qmix \\
roma\_raw & False \\
soft\_constraint\_weight & 1.0 \\
var\_floor & 0.002 \\
\hline
\end{tabular}
\caption{ROMA algorithm specific parameters}
\label{table:roma_parameters}
\end{table}

\begin{table}[htbp]
\centering
\begin{tabular}{|p{0.5\columnwidth}|p{0.4\columnwidth}|}
\hline
Parameter & Value \\
\hline
agent & rnn \\
central\_action\_embed & 1 \\
central\_agent & central\_rnn \\
central\_loss & 1 \\
central\_mac & basic\_central\_mac \\
central\_mixer & ff \\
central\_mixing\_embed\_dim & 128 \\
central\_rnn\_hidden\_dim & 64 \\
condition & max\_action \\
condition\_loss & mse \\
condition\_loss\_delta & 0.001 \\
double\_q & True \\
hypernet\_embed & 64 \\
hypernet\_layers & 2 \\
hysteretic\_qmix & False \\
learner & restq\_learner \\
mac & basic\_mac \\
max\_second\_gap & 0 \\
mixer & qmix \\
noopt\_loss & 1 \\
qmix\_loss & 1 \\
residual\_negative\_abs & True \\
resq\_version & v3 \\
run & default \\
td\_lambda & 0.6 \\
\hline
\end{tabular}
\caption{ResQ algorithm specific parameters}
\label{table:resq_parameters}
\end{table}

\begin{table}[htbp]
\centering
\begin{tabular}{|p{0.5\columnwidth}|p{0.4\columnwidth}|}
\hline
Parameter & Value \\
\hline
agent & iqn\_rnn \\
agent\_own\_state\_size & True \\
central\_agent & anyway \\
central\_loss & 1 \\
central\_mac & base\_central\_mac \\
central\_mixer & ff \\
central\_mixing\_embed\_dim & 128 \\
double\_q & True \\
hypernet\_embed & 64 \\
hypernet\_layers & 2 \\
learner & dresq \\
mac & basic\_mac \\
mixer & datten \\
n\_approx\_quantiles & 32 \\
n\_attention\_head & 4 \\
n\_constrant\_value & 32 \\
n\_head\_embedding\_layer1 & 64 \\
n\_head\_embedding\_layer2 & 4 \\
n\_key\_embedding\_layer1 & 32 \\
n\_quantiles & 8 \\
n\_query\_embedding\_layer1 & 64 \\
n\_query\_embedding\_layer2 & 32 \\
n\_target\_quantiles & 8 \\
negative\_abs & True \\
optimizer & RMSProp \\
qmix\_loss & 1 \\
quantile\_embed\_dim & 64 \\
td\_lambda & 0.6 \\
type & weighted \\
\hline
\end{tabular}
\caption{ResZ algorithm specific parameters}
\label{table:resz_parameters}
\end{table}

\begin{table}[htbp]
\centering
\begin{tabular}{|p{0.5\columnwidth}|p{0.4\columnwidth}|}
\hline
Parameter & Value \\
\hline
agent & rnn \\
critic\_baseline\_fn & coma \\
critic\_q\_fn & coma \\
critic\_train\_mode & seq \\
critic\_train\_reps & 1 \\
learner & offpg\_learner \\
mac & basic\_mac \\
mask\_before\_softmax & False \\
name & s\_tape \\
off\_batch\_size & 128 \\
off\_buffer\_size & 5000 \\
p & 0.3 \\
q\_nstep & 0 \\
step & 5 \\
tb\_lambda & 0.93 \\
td\_lambda & 0.8 \\
\hline
\end{tabular}
\caption{sTAPE algorithm specific parameters}
\label{table:stape_parameters}
\end{table}

\begin{table}[htbp]
\centering
\begin{tabular}{|p{0.5\columnwidth}|p{0.4\columnwidth}|}
\hline
Parameter & Value \\
\hline
agent & rnn \\
alpha\_init & -0.07 \\
alpha\_lr & 3e-4 \\
c\_beta & 1.0 \\
central\_action\_embed & 1 \\
central\_agent & central\_rnn \\
central\_loss & 1 \\
central\_mac & basic\_central\_mac \\
central\_mixer & ff \\
central\_mixing\_embed\_dim & 256 \\
central\_rnn\_hidden\_dim & 64 \\
comm & True \\
comm\_beta & 0.001 \\
comm\_beta\_end\_decay & 50000000 \\
comm\_beta\_start\_decay & 20000000 \\
comm\_beta\_target & 1e-2 \\
comm\_embed\_dim & 3 \\
comm\_entropy\_beta & 1e-6 \\
comm\_entropy\_beta\_end\_decay & 50000000 \\
comm\_entropy\_beta\_start\_decay & 20000000 \\
comm\_entropy\_beta\_target & 1e-4 \\
comm\_method & information\_bottleneck \\
critic\_agent & rnn\_agent\_n \\
critic\_mac & cate\_broadcast\_mac \\
cut\_mu\_rank\_thres & 80.0 \\
cut\_mu\_thres & 1.0 \\
double\_q & True \\
gate\_loss\_beta & 1e-5 \\
hypernet\_embed & 64 \\
hypernet\_layers & 2 \\
hysteretic\_qmix & True \\
is\_comm\_beta\_decay & False \\
is\_comm\_entropy\_beta\_decay & False \\
is\_cur\_mu & False \\
is\_print & False \\
is\_rank\_cut\_mu & False \\
learner & max\_q\_learner \\
mac & basic\_mac\_logic \\
mixer & qmix \\
name & ow\_qmix\_env=4\_dtape \\
only\_downstream & False \\
p & 0.5 \\
qmix\_loss & 1 \\
run & default \\
td\_lambda & 0.6 \\
use\_IB & True \\
w & 0.5 \\
\hline
\end{tabular}
\caption{dTAPE algorithm specific parameters}
\label{table:dtape_parameters}
\end{table}

\subsubsection{About LLM-PySC2 framework} Currently, we focus solely on agents' ability to interpret high-level strategic concepts within the HLSMAC scenarios by using stratagem names or brief explanations as prompts. The framework also supports more detailed prompts for further study. For each scenario, we set up the prompts about scenario-objective as follows.
\begin{enumerate}
    \item \textbf{adcc}: ``Use Openly repair the gallery roads, but sneak through the passage of Chencang to attack the enemy base at top right of the map.''
    \item \textbf{dhls}: ``Use Lure the Tiger Out of the Mountains to get the enemy army away from the enemy base near minimap position [40, 15], then attack the enemy base.''
    \item \textbf{fkwz}: ``The enemy is trying to spawn in an overwhelming army at the bottom left of the map, Use Make the Host and the Guest Exchange to stop it.''
    \item \textbf{gmzz}: ``Use the strategy Shut The Door to Catch the Thief to defeat the zerg army.''
    \item \textbf{jctq}: ``Use Slough Off the Cicada's Shell to escape.''
    \item \textbf{jdsr}: ``Use Kill With a Borrowed Knife to defeat the enemy army.''
    \item \textbf{sdjx}: ``Coordinate with your team and use Make a Feint to the East While Attacking in the West to distract the enemy.''
    \item \textbf{swct}: ``Use Pull Down the Ladder After the Ascent to attack the enemy base near the bottom right of the map.''
    \item \textbf{tlhz}: ``Use Replace the beams with rotten timbers to destroy the enemy buildings.''
    \item \textbf{wwjz}: ``There is an enemy army attacking your base. Use Besiege Wei to Rescue Zhao to defeat them. The enemy base is at the bottom right of the map.''
    \item \textbf{wzsy}: ``Use Create something from nothing to attack the enemy base.''
    \item \textbf{yqgz}: ``Use In order to capture, one must let loose to defeat the enemy.''
\end{enumerate}

\subsection{Detailed Metrics and Experimental Data}

\subsubsection{Detailed HLSMAC metrics} 
This section presents detailed metric definitions for evaluating agent performance for each HLSMAC scenario. The metrics are generally defined in the main text. It is important to note that not all scenarios incorporate all five metrics, as each scenario has its own unique objectives and unit compositions.

\begin{enumerate}
\item \textbf{Metrics: adcc (Secretly March to Chencang)}
\begin{itemize}
    \item \texttt{TPF}: The average frequency of allied Zerglings entering within a 6-unit radius around the enemy Command Center per game episode.    
    \item \texttt{TDA}: The projection ratio of each Zergling's actual displacement vector onto the shortcut vector from the spawning point to the enemy Command Center.
    \item \texttt{AUF}: The frequency of Zerglings casting \texttt{Burrow} and \texttt{Unburrow} abilities.
    \item \texttt{CTD}: The total damage dealt by allied units to the enemy Command Center.
    \item \texttt{USR}: The average ratio of surviving allied Zerglings to the initial count.
\end{itemize}

\item \textbf{Metrics: dhls (Lure the Tiger Down the Mountain)}
\begin{itemize}
    \item \texttt{TPF}: The average frequency of allied Zerglings and Roaches entering within a 6-unit radius around the enemy Command Center per game episode.
    \item \texttt{TDA}: The projection ratio of each Zergling's or Roach's actual displacement vector onto the shortcut vector from the spawning point to the enemy Command Center.
    \item \texttt{AUF}: The frequency of Nydus Networks and Nydus Worms casting \texttt{Load}/\texttt{Unload} abilities.
    \item \texttt{CTD}: The total damage dealt by allied Zerglings and Roaches to the enemy Command Center.
    \item \texttt{USR}: The average ratio of surviving allied Zerglings and Roaches to the initial count.
\end{itemize}

\item \textbf{Metrics: fkwz (Exchange the Role of Guest for That of Host)}
\begin{itemize}
    \item \texttt{TPF}: The average frequency of allied Warp Prism and Zealots entering within a 6-unit radius around the enemy Gateway per game episode.
    \item \texttt{TDA}: The projection ratio of each Warp Prism or Zealot's actual displacement vector onto the shortcut vector from the spawning point to the enemy Gateway.
    \item \texttt{AUF}: The frequency of allied Warp Prism casting \texttt{Load}/\texttt{Unload} and \texttt{Phasing Mode}, and the Warp Gates' \texttt{Warp In} abilities.
    \item \texttt{CTD}: The total damage dealt by allied Zealots to the enemy Gateway.
    \item \texttt{USR}: The average ratio of surviving allied Warp Prism to the initial count.
\end{itemize}

\item \textbf{Metrics: gmzz (Shut the Door to Catch the Thief)}
\begin{itemize}
    \item \texttt{TPF}: The average frequency of enemy Zerglings entering within a 6-unit radius around the centroid of the three allied Supply Depots per game episode.
    \item \texttt{AUF}: The frequency of Supply Depots casting \texttt{SupplyDepotLower} and \texttt{SupplyDepotRaise} abilities.
    \item \texttt{CTD}: The total damage dealt by allied Marines to enemy Zerglings.
    \item \texttt{USR}: The average ratio of allied Marines to the initial count.
\end{itemize}

\item \textbf{Metrics: jctq (Shed Your Skin Like the Golden Cicada)}
\begin{itemize}
    \item \texttt{TPF}: The average frequency of allied Roaches entering within a 6-unit radius around a target coordinate (e.g., (35,35)) per game episode.
    \item \texttt{TDA}: The projection ratio of each Roach's actual displacement vector onto the shortcut vector from the spawning point to coordinate (35,35).
    \item \texttt{AUF}: The frequency of Roaches casting \texttt{Burrow} abilities.
    \item \texttt{USR}: The average ratio of allied Roaches to the initial count.
\end{itemize}

\item \textbf{Metrics: jdsr (Kill with a Borrowed Sword)}
\begin{itemize}
    \item \texttt{AUF}: The frequency of Infestors casting \texttt{Neural Parasite} abilities.
    \item \texttt{CTD}: The total damage dealt by Roaches to enemy Stalkers and Colossi.
    \item \texttt{USR}: The average ratio of allied Roaches and Infestor to the initial count.
\end{itemize}

\item \textbf{Metrics: sdjx (Clamour in the East, Attack in the West)}
\begin{itemize}
    \item \texttt{TPF}: The average frequency of allied Medivacs entering within a 6-unit radius around the enemy main base Nexus per game episode.
    \item \texttt{TDA}: The projection ratio of each Medivac's actual displacement vector onto the shortcut vector from the spawning point to the enemy main base Nexus.
    \item \texttt{CTD}: The total damage dealt by allied units to the enemy expansion Nexus.
    \item \texttt{USR}: The average ratio of allied Medivacs and Marines to the initial count.
\end{itemize}

\item \textbf{Metrics: swct (Lure Your Enemy onto the Roof, then Take Away the Ladder)}
\begin{itemize}
    \item \texttt{TPF}: The average frequency of allied Sentries and Warp Prism entering within a 6-unit radius around the enemy Hatchery per game episode.
    \item \texttt{TDA}: The projection ratio of each allied Sentry or Warp Prism's actual displacement vector onto the shortcut vector from the spawning point to the enemy Hatchery.
    \item \texttt{AUF}: The frequency of Warp Prism casting \texttt{Load}/\texttt{Unload} abilities and Sentries casting \texttt{Force Field} abilities.
    \item \texttt{CTD}: The total damage dealt by allied Sentries and Warp Prism to the enemy Hatchery.
    \item \texttt{USR}: The average ratio of allied Sentries and Warp Prism to the initial count.
\end{itemize}

\item \textbf{Metrics: tlhz (Replace the Beams with Rotten Timbers)}
\begin{itemize}
    \item \texttt{TPF}: The average frequency of allied Evolution Chambers entering within a 6-unit radius around the enemy Photon Cannon per game episode.
    \item \texttt{AUF}: The frequency of Drones casting \texttt{Mutate into Hatchery}, \texttt{Cancel Building}, and \texttt{Mutate into Evolution Chamber} abilities, and Larvas casting \texttt{Morph to Zergling} abilities.
    \item \texttt{CTD}: The total damage dealt by allied units to enemy Photon Cannon and Pylon.
    \item \texttt{USR}: The average ratio of allied Drone and Larvas to the initial count.
\end{itemize}

\item \textbf{Metrics: wwjz (Besiege Wei to Rescue Zhao)}
\begin{itemize}
    \item \texttt{TPF}: The average frequency of allied Zealots entering within a 6-unit radius around the enemy Command Center per game episode.
    \item \texttt{TDA}: The projection ratio of each Zealot's actual displacement vector onto the shortcut vector from the spawning point to the enemy Command Center.
    \item \texttt{CTD}: The total damage dealt by allied Zealots to the enemy Command Center.
    \item \texttt{USR}: The average ratio of allied Zealots to the initial count.
\end{itemize}

\item \textbf{Metrics: wzsy (Create Something from Nothing)}
\begin{itemize}
    \item \texttt{TPF}: The average frequency of allied Stalkers and Sentries entering within a 6-unit radius around the enemy Pylon per game episode.
    \item \texttt{TDA}: The projection ratio of each allied unit's actual displacement vector onto the shortcut vector from the spawning point to the enemy Pylon.
    \item \texttt{AUF}: The frequency of Sentries casting \texttt{Hallucination} abilities.
    \item \texttt{CTD}: The total damage dealt by allied units to the enemy Pylon.
    \item \texttt{USR}: The average ratio of allied Stalkers and Sentries to the initial count.
\end{itemize}

\item \textbf{Metrics: yqgz (To Catch Something, First Let It Go)}
\begin{itemize}
    \item \texttt{TPF}: The average frequency of allied Zerglings entering within a 6-unit radius around the enemy Siege Tanks per game episode.
    \item \texttt{TDA}: The projection ratio of each allied Zergling's actual displacement vector onto the shortcut vector from the spawning point to the initial center of the enemy forces' formation.
    \item \texttt{CTD}: The total damage dealt by allied Zerglings to the enemy Marines and Siege Tanks.
    \item \texttt{USR}: The average ratio of allied Zerglings to the initial count.
\end{itemize}
\end{enumerate}

\subsubsection{Results of MARL baselines}
\begin{figure*}[htbp]
\centering
\includegraphics[width=2\columnwidth]{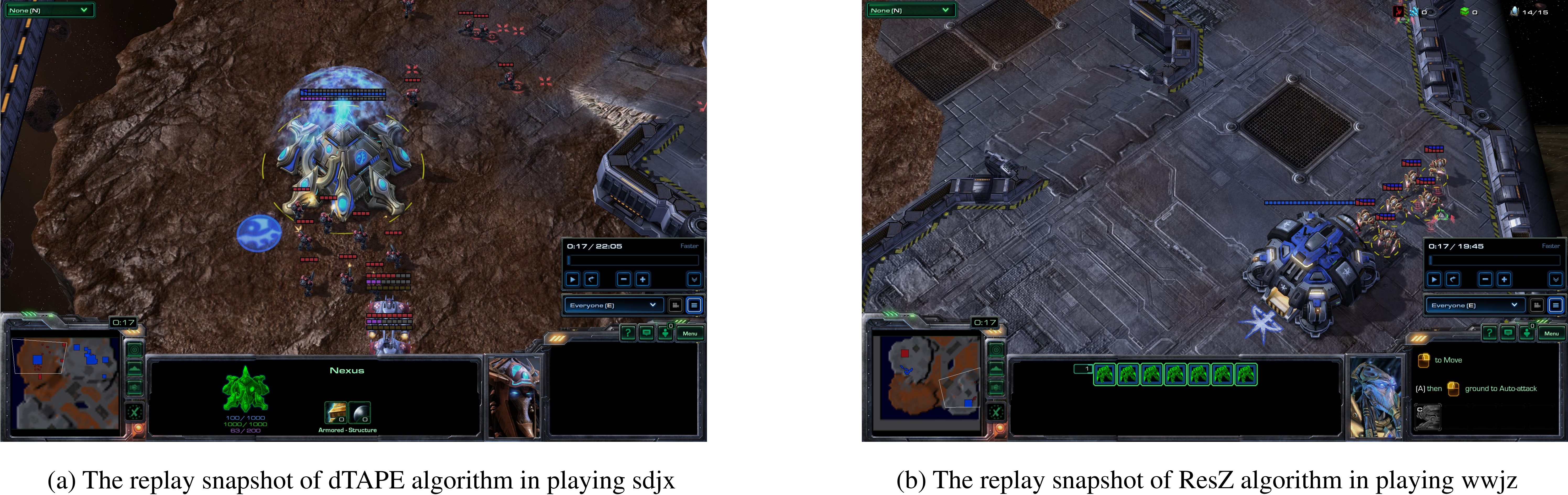} 
\caption{Replay snapshots of dTAPE on \textbf{sdjx} and RESZ on \textbf{wwjz}. In the left, medivacs execute a feint attack on the enemy main base while the allied main force simultaneously attacks the expansion base. In the right, allied forces advance along the map's right edge, forcing enemies to abandon their siege and retreat to defend their main base.}
\label{fig:sdjx_wwjz_example}
\end{figure*}

\begin{figure*}[htbp]
\centering
\includegraphics[width=2\columnwidth]{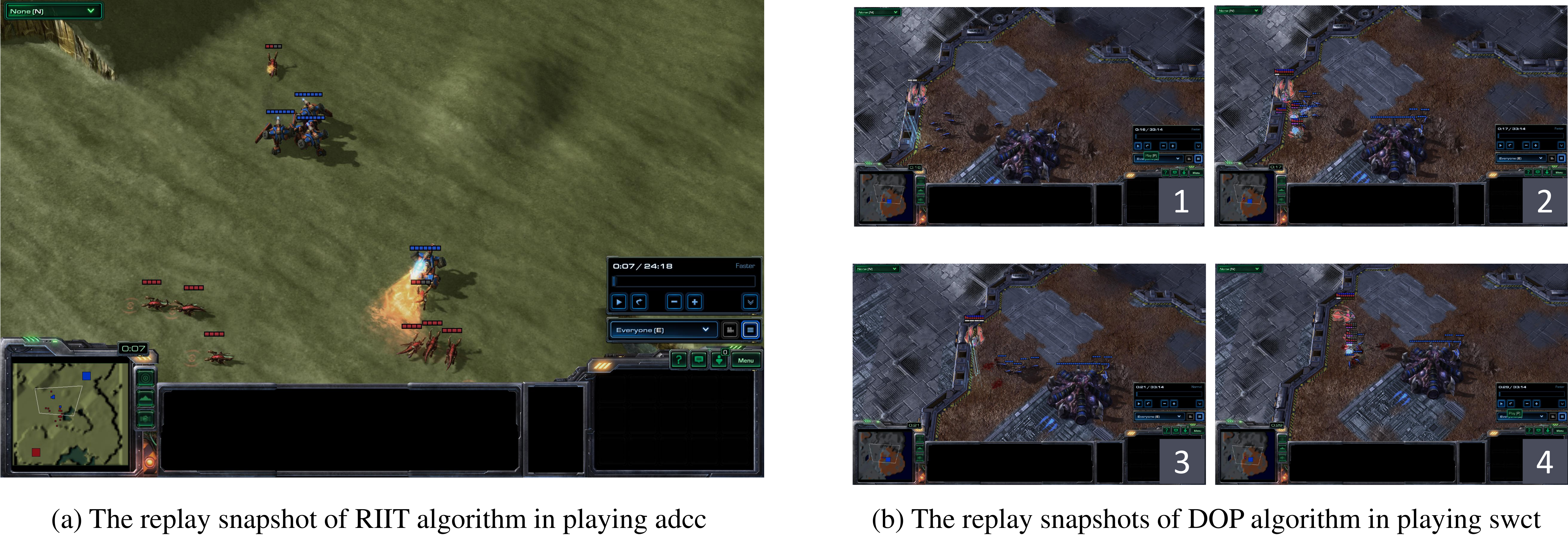} 
\caption{Replay snapshots of RIIT on \textbf{adcc} and DOP on \textbf{swct}. In the left, the allied Zerglings do not employ the ``secretly march to Chencang'' stratagem. Instead, they engage directly and split the units: deploying a few Zerglings as decoys to distract most Hellbats while the majority focus fire on individual targets before advancing to the enemy Command Center. In the right, the allied Warp Prism repeatedly picks up and drops off Sentries to keep them safe while executing hit-and-run tactics.}
\label{fig:riit_and_dop}
\end{figure*}

\begin{figure*}[htbp]
\centering
\includegraphics[width=2\columnwidth]{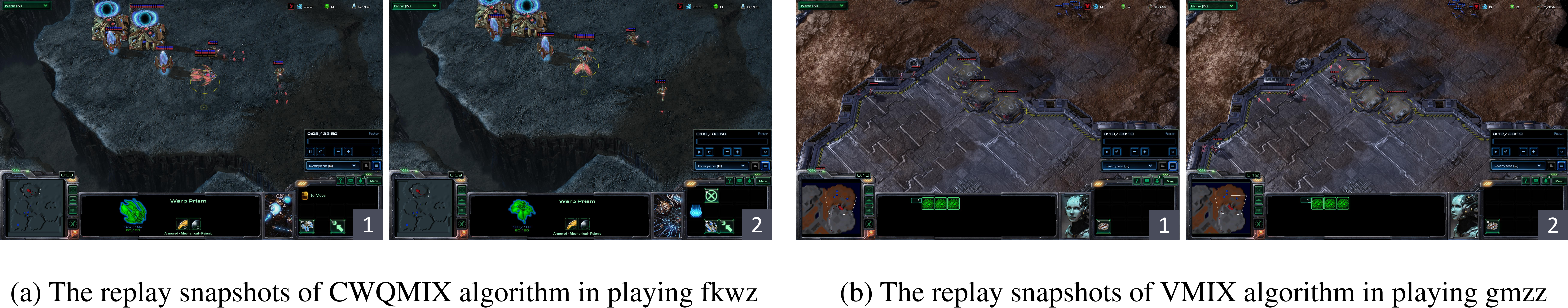} 
\caption{Replay snapshots of CWQMIX on \textbf{fkwz} and VMIX on \textbf{gmzz}. In the left, warp prism repeatedly switches between phasing and transport modes in fkwz. In the right, MARL methods like VMIX execute abilities hundreds of times aimlessly, while human experts use \texttt{SupplyDepotLower} and \texttt{SupplyDepotRaise} only once each to strategically block enemy Zerglings.}
\label{fig:fkwz_gmzz_AUF}
\end{figure*}

We evaluate 21 MARL algorithms from open-source repositories using default parameters. HLSMAC poses significant challenges for both MARL. Nearly $80\%$ of algorithm-scenario combinations achieve zero win rates as shown in Table~\ref{tab:marl-win-rates}. Figures~\ref{fig:sdjx_wwjz_example}-\ref{fig:fkwz_gmzz_AUF} exemplify the diverse behaviors observed in current MARL methods, ranging from `stratagem-aligned execution' to aimless repetitive actions.

Traditional win-rate metrics inadequately capture agents' human-like strategic decision-making capabilities in HLSMAC tasks. For example, RIIT achieves a $93\%$ win rate in \textit{adcc}, but replay analysis reveals it does not actually follow the intended stratagem approach. Similarly, DOP achieves only a $19\%$ win rate in \textit{swct}, yet exhibits repeated loading and unloading of Sentries with Warp Prism, a behavior rarely seen in normal human play. In addition, high win rates in \textit{jdsr} do not guarantee that the Infester actually comprehends the ``borrow sword'' concept, even when casting \texttt{Neural Parasite}.

Our five new metrics, together with win rate, offer richer dimensions for evaluating the performance of various methods. Through $R^2$ analysis of valid metric-scenario combinations, we find that these metrics demonstrate strong explanatory power. Metrics such as TPF, CTD, and TDA, as well as USR, serve as effective indicators of high win-rate performance in the \textit{wzsy}, \textit{wwjz}, and \textit{sdjx} scenarios. This reflects a fundamental strategic truth: success depends on approaching and neutralizing critical targets, which the \textit{wwjz} and \textit{sdjx} scenarios clearly exemplify. Furthermore, although Ability Utilization Frequency (AUF) has a weaker correlation with win rate, likely because this frequency-based metric cannot fully reflect the appropriateness of ability usage, it proves valuable for distinguishing human players from current methods, as it reveals that humans demonstrate purposeful utilization of critical abilities whereas current methods do not (see Table~\ref{tab:auf}). As shown in Figure~\ref{fig:fkwz_gmzz_AUF}.

\begin{table*}[t]
\centering
\begin{tabular}{ccccccccccccc}
\hline
& \textbf{adcc} & \textbf{dhls} & \textbf{fkwz} & \textbf{gmzz} & \textbf{jctq} & \textbf{jdsr} & \textbf{sdjx} & \textbf{swct} & \textbf{tlhz} & \textbf{wwjz} & \textbf{wzsy} & \textbf{yqgz} \\ \hline
IQL    & 0.3 & 0 & 0 & 0.04 & 0 & 0.4 & 0 & 0.03 & 0 & 0 & 0.34 & 0 \\
COMA   & 0 & 0 & 0 & 0 & 0 & 0.02 & 0 & 0 & 0 & 0 & 0 & 0 \\
VDN    & 0 & 0 & 0 & 0.12 & 0 & 0.6 & 0.14 & 0 & 0 & 0 & 0.09 & 0 \\
QMIX   & 0 & 0 & 0 & 0.01 & 0 & 0.88 & 0 & 0 & 0 & 0 & 0.85 & 0 \\
VMIX   & 0 & 0 & 0 & 0 & 0 & 0 & 0 & 0 & 0 & 0 & 0 & 0 \\
MAVEN  & 0 & 0 & 0 & 0 & 0 & 0.74 & 0 & 0 & 0 & 0 & 0 & 0 \\
QTRAN  & 0 & 0 & 0 & 0.12 & 0 & 0.82 & 0.32 & 0 & 0 & 0 & 0 & 0 \\
CWQMIX & 0 & 0 & 0 & 0 & 0 & 0.42 & 0.83 & 0 & 0 & 0.01 & 0.96 & 0 \\
OWQMIX & 0 & 0 & 0 & 0 & 0 & 0.85 & 0 & 0 & 0 & 0 & 0 & 0 \\
DOP    & 0 & 0 & 0 & 0.26 & 0 & 1 & 0 & 0.19 & 0 & 0 & 0 & 0 \\
LICA   & 0 & 0 & 0 & 0.5 & 0 & 0.73 & 0 & 0 & 0 & 0 & 0.02 & 0 \\
Qatten & 0 & 0 & 0 & 0.15 & 0 & 0.92 & 0.88 & 0 & 0 & 0 & 0.94 & 0 \\
QPLEX  & 0 & 0 & 0 & 0 & 0 & 0.93 & 0.86 & 0 & 0 & 0 & 0 & 0 \\
FOP    & 0 & 0 & 0 & 0 & 0.03 & 0.73 & 0 & 0 & 0 & 0 & 0 & 0 \\
RIIT   & 0.93 & 0 & 0 & 0.27 & 0 & 0.96 & 0 & 0 & 0 & 0 & 0 & 0 \\
RODE   & 0 & 0 & 0 & 0 & 0 & 0 & 0 & 0 & 0 & 0 & 0 & 0 \\
ROMA   & 0 & 0 & 0 & 0 & 0 & 0.77 & 0 & 0 & 0 & 0 & 0 & 0 \\
RESQ   & 0 & 0 & 0 & 0.41 & 0 & 0.91 & 0 & 0 & 0 & 0 & 0 & 0 \\
RESZ   & 0 & 0 & 0 & 0.26 & 0 & 0.79 & 0.72 & 0 & 0 & 0.78 & 0.07 & 0 \\
dTAPE  & 0.49 & 0 & 0 & 0.84 & 0 & 0.93 & 0.89 & 0 & 0 & 1 & 1 & 0 \\
sTAPE  & 0 & 0 & 0 & 0 & 0 & 0.98 & 0 & 0 & 0 & 0 & 0 & 0 \\ \hline
GPT-3.5 Example  & 0 & 0 & 0 & 0 & 0 & 0 & 0 & 0 & 0 & 0 & 0 & 0 \\ \hline
Human Example  & 1 & 1 & 1 & 1 & 1 & 1 & 1 & 1 & 1 & 1 & 1 & 1 \\ \hline
\end{tabular}%
\caption{Win Rates Across Diverse Methods on the 12 HLSMAC Scenarios}
\label{tab:marl-win-rates}
\end{table*}

\begin{figure*}[htbp]
\centering
\includegraphics[width=2\columnwidth]{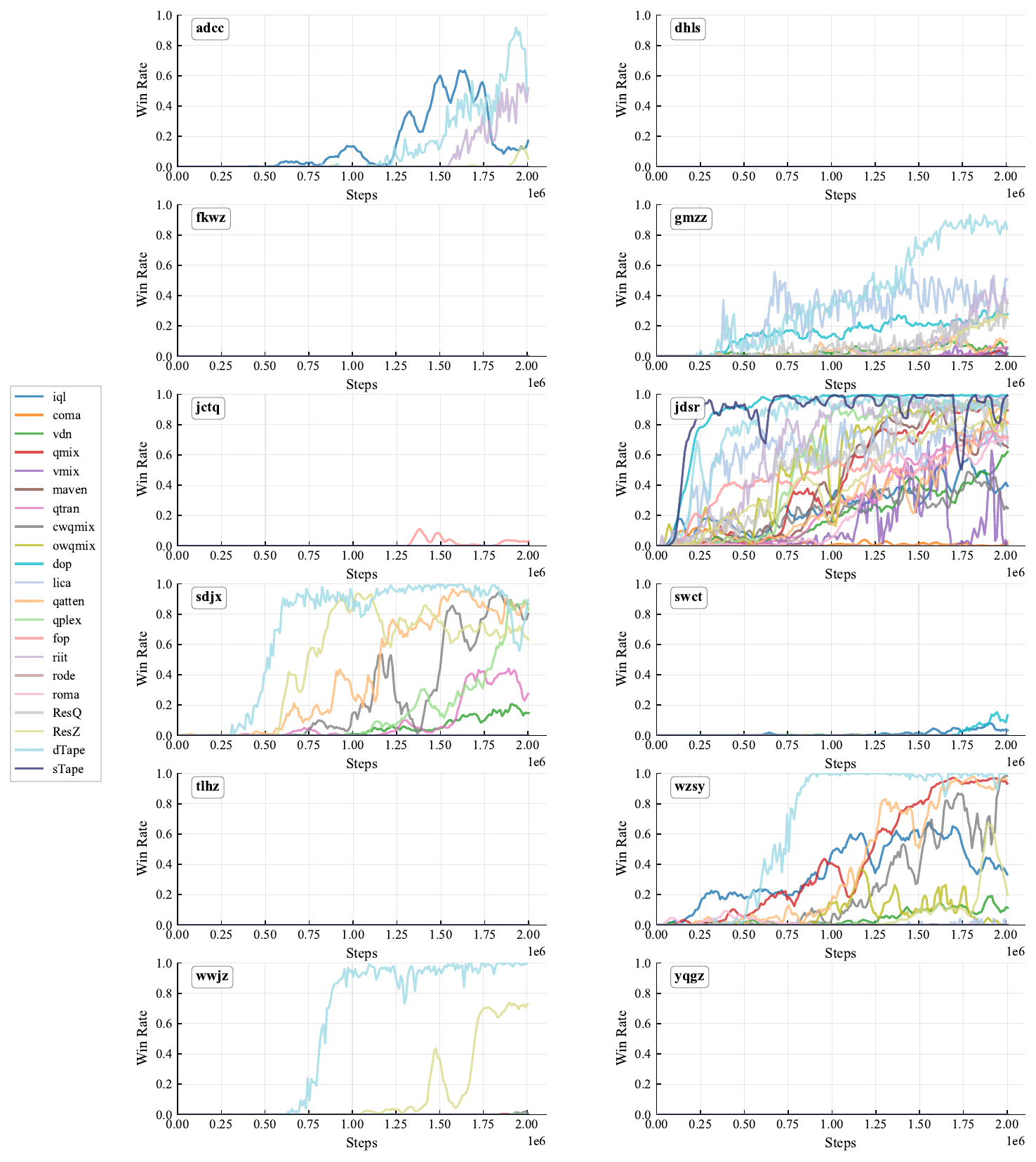} 
\caption{Win Rates Across 21 MARL Baselines on the 12 HLSMAC Scenarios. On the scenarios \textbf{dhls}, \textbf{fkwz},\textbf{tlhz} and \textbf{yqgz}, all algorithms achieve zero win rates. Some figures show fewer curves because many algorithms achieve consistently poor performance (zero or near-zero win rates) during training.}
\label{fig:marl-win-rates}
\end{figure*}

\begin{table*}[t]
\centering
\begin{tabular}{ccccccccccccc}
\hline
& \textbf{adcc} & \textbf{dhls} & \textbf{fkwz} & \textbf{gmzz} & \textbf{jctq} & \textbf{jdsr} & \textbf{sdjx} & \textbf{swct} & \textbf{tlhz} & \textbf{wwjz} & \textbf{wzsy} & \textbf{yqgz} \\ \hline
IQL    & 167.66 & 0 & 0 & 103.41 & 18.33 & --    & 0  & 0   & 0 & 0   & 50  & 0 \\
COMA   & 0   & 0 & 0 & 0   & 0.25  & --    & 0  & 0   & 0 & 0   & 0   & 0 \\
VDN    & 0   & 0 & 0 & 116.95 & 21.4 & --    & 0  & 0   & 0 & 175.96 & 22.3  & 0 \\
QMIX   & 0   & 0 & 0 & 0   & 25.8 & --    & 0  & 0   & 0 & 122.28 & 80.4  & 0 \\
VMIX   & 0.05   & 0 & 0 & 0   & 1.61  & --    & 1.25  & 0   & 0 & 16.51  & 0   & 0 \\
MAVEN  & 0   & 0 & 0 & 150.21 & 34.81 & --    & 0  & 0   & 0 & 63.32  & 22.52  & 0 \\
QTRAN  & 0   & 0 & 0 & 107.96 & 63.47 & --    & 5.44  & 0   & 0 & 15.69  & 0   & 0 \\
CWQMIX & 0   & 0 & 0 & 0   & 23.88 & --    & 8.41  & 0   & 0 & 148.63 & 86.09  & 0 \\
OWQMIX & 0   & 0 & 0 & 0   & 14.3 & --    & 0  & 0   & 0 & 0   & 0   & 0 \\
DOP    & 0   & 0 & 0 & 0   & 2.64  & --    & 0  & 168.87 & 0 & 0   & 0   & 0 \\
LICA   & 0   & 0 & 0.32 & 131.88 & 6.04  & --    & 0  & 0   & 0 & 89.63  & 0.92   & 0 \\
Qatten & 0   & 0 & 0 & 134.11 & 21.51 & --    & 13.19 & 0   & 0 & 129.38 & 104.8 & 0 \\
QPLEX  & 0.25   & 0 & 0 & 857.71 & 30.01 & --    & 11.06 & 0   & 0 & 195.64 & 0   & 0 \\
FOP    & 0   & 0 & 0 & 0   & 0.4  & --    & 0  & 0   & 0 & 0   & 0   & 0 \\
RIIT   & 90.58  & 0 & 0 & 0   & 2.07  & --    & 0  & 0   & 0 & 0   & 4.65   & 0 \\
RODE   & 0   & 0 & 0 & 0   & 26.1 & --    & 0  & 0   & 0 & 74.33  & 0   & 0 \\
ROMA   & 0   & 0 & 0 & 164.32 & 18.91 & --    & 0  & 9.23   & 0 & 69.14  & 0   & 0 \\
RESQ   & 0   & 0 & 0 & 119.02 & 24.04 & --    & 0  & 0   & 0 & 0   & 0   & 0 \\
RESZ   & 0   & 0 & 0 & 136.71 & 34.75 & --    & 4.31  & 0   & 0 & 368.07 & 7.73   & 0 \\
dTAPE  & 252.57 & 0 & 0 & 137.31 & 36.54 & --    & 8.7  & 0   & 0 & 338.7 & 83.61  & 0 \\
sTAPE  & 0   & 0 & 0 & 0   & 8.27  & --    & 0  & 0   & 0 & 0   & 0   & 0 \\ \hline
GPT-3.5 Example& 0 &	0 &0	&0&	0	&	-- &0	&209.2	&0&	0&	0	&0 \\ \hline
Human Example & 330.69 &	202.15&	238.2&	155.38&	0&	--&	0&	588.8&	0&	225.71&	88.83&	0 \\ \hline
\end{tabular}
\caption{Target Proximity Frequency (TPF) Across Diverse Methods on the 12 HLSMAC Scenarios}
\label{tab:tpf}
\end{table*}

\begin{table*}[t] % [t] places the table at the top of the page
\centering
\begin{tabular}{ccccccccccccc}
\hline
& \textbf{adcc} & \textbf{dhls} & \textbf{fkwz} & \textbf{gmzz} & \textbf{jctq} & \textbf{jdsr} & \textbf{sdjx} & \textbf{swct} & \textbf{tlhz} & \textbf{wwjz} & \textbf{wzsy} & \textbf{yqgz} \\ \hline
IQL    & 0.69 & 0.17 & 0.67  & -- & -0.05 & -- & -0.97 & -0.5 & -- & 0.03 & 0.7 & 0.32 \\
COMA   & 0.15 & 0.98 & 0.69  & -- & 0.72  & -- & -0.95 & -0.33 & -- & 0.64 & 0.47 & 0.13 \\
VDN    & 0.39 & 0.64 & 0.34  & -- & -0.07 & -- & 0.94  & -0.61 & -- & 0.93 & 0.64 & 0.29 \\
QMIX   & 0.39 & 0.07 & 0.36  & -- & -0.1 & -- & -0.33 & 0.1  & -- & 0.94 & 0.96 & 0.29 \\
VMIX   & 0.26 & 0.37 & 0.72  & -- & 0.61  & -- & 0.63  & -0.57 & -- & 0.86 & 0.68 & 0.2 \\
MAVEN  & 0.27 & 0.49 & 0.08  & -- & 0.08  & -- & 0.14  & 0.39  & -- & 0.91 & 0.56 & 0.44 \\
QTRAN  & 0.35 & 0.98 & -0.02 & -- & 0.46  & -- & 0.78  & -0.57 & -- & 0.29 & 0.4 & 0.02 \\
CWQMIX & 0.87 & 0.98 & 0.49  & -- & -0.06 & -- & 0.93  & -0.23 & -- & 0.95 & 0.95 & 0.1 \\
OWQMIX & 0.48 & 0.97 & 0.39  & -- & -0.17 & -- & -0.95 & 0.54  & -- & 0.38 & 0.83 & 0.01 \\
DOP    & 0.42 & 0.5 & 0.71  & -- & 0.78  & -- & -0.95 & 0.88  & -- & 0.06 & 0.76 & 0.01 \\
LICA   & 0.92 & 0.53 & 0.29  & -- &0& -- & 0.7  & 0.46  & -- & 0.66 & 0.6 & 0.37 \\
Qatten & 0.42 & 0.49 & 0.49  & -- & -0.1 & -- & 0.96  & -0.39 & -- & 0.95 & 0.87 & 0.28 \\
QPLEX  & 0.31 & 0.38 & 0.11  & -- &0& -- & 0.98  & -0.45 & -- & 0.97 & 0.49 & 0.42 \\
FOP    & 0.7 & 0.84 & 0.21  & -- & 0.62  & -- & -0.33 &0& -- & 0.41 & 0.93 & 0.11 \\
RIIT   & 0.94 & 0.92 & 0.24  & -- & -0.1 & -- & 0.1  & 0.01  & -- & 0.71 & 0.69 & 0.04 \\
RODE   & 0.91 & 0.59 & 0.52  & -- & -0.06 & -- & -0.95 & -0.67 & -- & 0.89 & 0.27 & 0.24 \\
ROMA   & 0.4 & 0.83 & 0.57  & -- & -0.12 & -- & -0.48 & 0.06  & -- & 0.93 & 0.4 & 0.34 \\
RESQ   & 0.99 & 0.71 & 0.57  & -- & -0.14 & --    & -0.95 & 0.6  & -- & 0.92 & 0.47 & 0.33 \\
RESZ   & 0.99 & 0.15 & 0.65  & -- & -0.05 & --    & 0.89  & -0.32 & --    & 1    & 0.63 & 0.09 \\
dTAPE  & 0.94 & 0.87 & 0.35  & -- & -0.06 & --    & 0.94  & -0.15 & --    & 1    & 0.76 & 0.03 \\
sTAPE  & 0.5 & 0.72 & 0.58  & -- & 0.81  & --    & -0.95 & -0.32 & --    & 0.5 & 0.25 & 0.13 \\ \hline
GPT-3.5 Example & 0.98 & 0 &	0.14&	-- &	-0.92&	--&	0.21&	0.5&	--&	0.76&	0.85&	0.02\\ \hline
Human Example & 1 &	1 &	0.99 &	-- &	-0.99&	--&	0 &	0.85&	--&	0.98&	0.99&	1 \\ \hline
\end{tabular}
\caption{Target Directional Alignment (TDA) Across Diverse Methods on the 12 HLSMAC Scenarios}
\label{tab:tda}
\end{table*}

\begin{table*}[t] % [t] places the table at the top of the page
\centering
\begin{tabular}{ccccccccccccc}
\hline
& \textbf{adcc} & \textbf{dhls} & \textbf{fkwz} & \textbf{gmzz} & \textbf{jctq} & \textbf{jdsr} & \textbf{sdjx} & \textbf{swct} & \textbf{tlhz} & \textbf{wwjz} & \textbf{wzsy} & \textbf{yqgz} \\ \hline
IQL    & 91.53  & 4.88  & 41.25  & 439.91 & 0 & 1    & -- & 21.47  & 8.53  & -- & 3 & -- \\
COMA   & 0 & 25.94 & 1.75   & 600 & 0 & 1    & -- & 0 & 4  & -- & 0.22 & -- \\
VDN    & 42.84  & 3.91  & 17.78  & 128.34 & 1.13 & 1    & -- & 5.5   & 6.13  & -- & 3 & -- \\
QMIX   & 37.78  & 0 & 75.16  & 0 & 0 & 1    & -- & 0 & 5.31  & -- & 3 & -- \\
VMIX   & 240.97 & 13 & 4.47   & 594.16 & 0 & 1    & -- & 200 & 5  & -- & 3 & -- \\
MAVEN  & 419.75 & 6.19  & 48.97  & 77.44  & 0.41 & 1    & -- & 69.16  & 2.53  & -- & 2.75 & -- \\
QTRAN  & 15.31  & 22.91 & 97.91  & 163.91 & 0 & 1    & -- & 200 & 5  & -- & 3 & -- \\
CWQMIX & 0.06   & 0 & 114.22 & 73.47  & 0.06 & 1    & -- & 4   & 4  & -- & 3 & -- \\
OWQMIX & 73.25  & 0 & 106.25 & 0 & 0 & 1    & -- & 0 & 6  & -- & 3 & -- \\
DOP    & 1.75   & 0 & 2.69   & 524.94 & 0 & 1    & -- & 49.63  & 2.06  & -- & 3 & -- \\
LICA   & 58.66  & 7.56  & 24.84  & 264.53 & 0 & 1    & -- & 3.94   & 1  & -- & 3 & -- \\
Qatten & 842.66 & 15.31 & 35.31  & 69.94  & 0.13 & 1    & -- & 109.66 & 11.5 & -- & 3 & -- \\
QPLEX  & 315 & 6.09  & 17.34  & 75.03  & 0.22 & 1    & -- & 176.41 & 3  & -- & 3 & -- \\
FOP    & 1.06   & 0 & 64.06  & 388 & 0 & 1    & -- & 4   & 7  & -- & 0 & -- \\
RIIT   & 25.41  & 20.38 & 77.09  & 389.47 & 2.41 & 1    & -- & 4.03   & 4  & -- & 3 & -- \\
RODE   & 74.34  & 8.47  & 3.97   & 0 & 0 & 1    & -- & 0 & 1  & -- & 3 & -- \\
ROMA   & 337.75 & 18.34 & 35.66  & 103.88 & 0.16 & 0.94 & -- & 11.84  & 7.72  & -- & 2.88 & -- \\
RESQ   & 228.75 & 0 & 10.28  & 20.88  & 0.28 & 1    & -- & 0 & 7  & -- & 3 & -- \\
RESZ   & 79.81  & 0 & 5.88   & 6   & 0 & 1    & -- & 0 & 4  & -- & 3 & -- \\
dTAPE  & 38.91  & 1.66  & 2.88   & 31.06  & 0 & 1    & -- & 1.53   & 7  & -- & 3 & -- \\
sTAPE  & 178.91 & 0 & 1.81   & 600 & 0 & 1    & -- & 0 & 9.28  & -- & 3 & -- \\ \hline
GPT-3.5 Example& 35 & 0 & 7 & 20& 3 &  0 & --  & 8 & 0 & -- & 0 & --\\ \hline
Human Example & 0/2 & 1 & 6 & 2 & 2& 1& -- & 7 & 6 & -- & 7 & -- \\ \hline
\end{tabular}
\caption{Ability Utilization Frequency (AUF) Across Diverse Methods on the 12 HLSMAC Scenarios}
\label{tab:auf}
\end{table*}

\begin{table*}[t] % [t] places the table at the top of the page
\centering
\begin{tabular}{ccccccccccccc}
\hline
& \textbf{adcc} & \textbf{dhls} & \textbf{fkwz} & \textbf{gmzz} & \textbf{jctq} & \textbf{jdsr} & \textbf{sdjx} & \textbf{swct} & \textbf{tlhz} & \textbf{wwjz} & \textbf{wzsy} & \textbf{yqgz} \\ \hline
IQL    & 0.48 & 0 & 0 & 0.28 & --    & 2.3 & 0 & 0 & 0 & 0 & 0.23 & 0 \\
COMA   & 0 & 0 & 0 & 0 & --    & 1.66 & 0 & 0 & 0 & 0 & 0 & 0 \\
VDN    & 0 & 0 & 0 & 0.41 & --    & 3.49 & 0 & 0 & 0 & 0.46 & 0.02 & 0 \\
QMIX   & 0 & 0 & 0 & 0 & --    & 3.5 & 0 & 0 & 0 & 0.3 & 0.91 & 0 \\
VMIX   & 0 & 0 & 0 & 0.07 & --    & 2.19 & 0 & 0 & 0 & 0.04 & 0 & 0 \\
MAVEN  & 0 & 0 & 0 & 0.39 & --    & 3.34 & 0 & 0 & 0 & 0.17 & 0 & 0 \\
QTRAN  & 0 & 0 & 0 & 0.33 & --    & 3.61 & 0.93 & 0 & 0 & 0.04 & 0 & 0 \\
CWQMIX & 0 & 0 & 0 & 0 & --    & 3.12 & 0.98 & 0 & 0 & 0.4 & 0.94 & 0 \\
OWQMIX & 0 & 0 & 0 & 0 & --    & 3.18 & 0 & 0 & 0 & 0 & 0 & 0 \\
DOP    & 0 & 0 & 0 & 0.61 & --    & 2.79 & 0 & 0.05 & 0 & 0 & 0 & 0 \\
LICA   & 0 & 0 & 0 & 0.6 & --    & 3.41 & 0 & 0 & 0 & 0.28 & 0 & 0 \\
Qatten & 0 & 0 & 0 & 0.51 & --    & 3.4 & 0.96 & 0 & 0 & 0.33 & 0.96 & 0 \\
QPLEX  & 0 & 0 & 0 & 0 & --    & 3.57 & 0.97 & 0 & 0 & 0.61 & 0 & 0 \\
FOP    & 0 & 0 & 0 & 0 & --    & 3.57 & 0 & 0 & 0 & 0 & 0 & 0 \\
RIIT   & 0.32 & 0 & 0 & 0.72 & --    & 3.27 & 0 & 0 & 0 & 0 & 0 & 0 \\
RODE   & 0 & 0 & 0 & 0 & --    & 1.96 & 0 & 0 & 0 & 0.07 & 0 & 0 \\
ROMA   & 0 & 0 & 0 & 0.27 & --    & 3.35 & 0 & 0 & 0 & 0.05 & 0 & 0 \\
RESQ   & 0 & 0 & 0 & 0.64 & --    & 3.33 & 0 & 0 & 0 & 0 & 0 & 0 \\
RESZ   & 0 & 0 & 0 & 0.85 & --    & 3.73 & 0.99 & 0 & 0 & 0.98 & 0.04 & 0 \\
dTAPE  & 0.73 & 0 & 0 & 0.89 & --    & 3.58 & 0.97 & 0 & 0 & 1    & 0.96 & 0 \\
sTAPE  & 0 & 0 & 0 & 0 & --    & 2.88 & 0 & 0 & 0 & 0 & 0 & 0 \\ \hline
% GPT-3.5 Example& 0 &	0 &0	&0&	0	&	-- &0	&209.2	&0&	0&	0	&0 \\ \hline
GPT-3.5 Example &0&	0 &0&	0&	--&	0.47&	0	&0.07&	0.08&	0&	0 & 0\\ \hline
Human Example & 1	& 1 &	0.09& 	0.88&	-- &	3.52&	1&	1&	0.98&	0.57&	0.97&	5.33 \\
\hline
\end{tabular}
\caption{{Critical Target Damage} (CTD) Across Diverse Methods on the 12 HLSMAC Scenarios}
\label{tab:ctd}
\end{table*}

\begin{table*}[t] % [t] places the table at the top of the page
\centering
\begin{tabular}{ccccccccccccc}
\hline
& \textbf{adcc} & \textbf{dhls} & \textbf{fkwz} & \textbf{gmzz} & \textbf{jctq} & \textbf{jdsr} & \textbf{sdjx} & \textbf{swct} & \textbf{tlhz} & \textbf{wwjz} & \textbf{wzsy} & \textbf{yqgz} \\ \hline
IQL    & 0.95 & 1 & 0.97 & 1 & 0.24 & 0.49 & 0.74 & 0.48 & 1 & 1    & 0.4 & 1 \\
COMA   & 1 & 1    & 0.03 & 1 & 0.38 & 0.33 & 1 & 1    & 0.98 & 1 & 0.41 & 1 \\
VDN    & 0.88 & 1 & 1    & 0.98 & 0.38 & 0.51 & 0.68 & 1 & 1    & 0.43 & 0.38 & 1 \\
QMIX   & 0.75 & 1 & 1    & 1 & 0.3 & 0.66 & 0.99 & 1 & 0.98 & 0.46 & 0.97 & 1 \\
VMIX   & 0.98 & 1 & 0.22 & 0.99 & 0.35 & 0.17 & 1 & 1    & 0.91 & 0.88 & 0.38 & 0.99 \\
MAVEN  & 0.97 & 1 & 1    & 0.98 & 0.31 & 0.7 & 0.74 & 0.98 & 0.91 & 0.54 & 0.36 & 0.95 \\
QTRAN  & 1 & 0.82 & 1 & 1    & 0.34 & 0.67 & 0.89 & 1 & 0.98 & 0.94 & 0.41 & 1 \\
CWQMIX & 1 & 1    & 1 & 1    & 0.31 & 0.43 & 0.9 & 1 & 0.97 & 0.53 & 0.98 & 1 \\
OWQMIX & 0.97 & 1 & 0.91 & 1 & 0.36 & 0.79 & 1 & 1    & 0.98 & 1 & 0.39 & 1 \\
DOP    & 0.97 & 1 & 0.09 & 1 & 0.44 & 0.99 & 1 & 0.61 & 1 & 1    & 0.38 & 1 \\
LICA   & 1 & 1    & 0.97 & 0.97 & 0.33 & 0.51 & 0.8 & 1 & 0.99 & 0.65 & 0.4 & 1 \\
Qatten & 0.78 & 1 & 1    & 0.98 & 0.37 & 0.73 & 0.92 & 0.94 & 1 & 0.47 & 0.99 & 1 \\
QPLEX  & 0.79 & 1 & 1    & 1 & 0.33 & 0.85 & 0.85 & 0.78 & 1 & 0.42 & 0.34 & 1 \\
FOP    & 1 & 1    & 1 & 1    & 0.4 & 0.51 & 1 & 1    & 1 & 1    & 0.44 & 1 \\
RIIT   & 0.87 & 1 & 0.97 & 1 & 0.38 & 0.85 & 0.8 & 1 & 0.97 & 1 & 0.4 & 1 \\
RODE   & 0.87 & 1 & 1    & 1 & 0.38 & 0.29 & 1 & 1    & 1 & 0.68 & 0.45 & 1 \\
ROMA   & 0.96 & 1 & 0.94 & 0.9 & 0.37 & 0.61 & 1 & 0.89 & 1 & 0.59 & 0.39 & 1 \\
RESQ   & 0.95 & 1 & 0.88 & 1 & 0.38 & 0.76 & 0.89 & 1 & 0.98 & 1 & 0.4 & 1 \\
RESZ   & 0.93 & 1 & 0.5 & 1 & 0.38 & 0.58 & 1 & 1    & 0.98 & 0.63 & 0.4 & 1 \\
dTAPE  & 0.96 & 0.96 & 0.31 & 1 & 0.39 & 0.88 & 0.97 & 1 & 0.98 & 1 & 1    & 1 \\
sTAPE  & 1 & 1    & 0.09 & 1 & 0.51 & 0.98 & 1 & 1    & 1 & 1    & 0.43 & 1 \\ \hline
GPT-3.5 Example& 0.88 &1 &	1&	1	& 1&	0.6	& 0.83&	0.2&	1&	1&	1&	1 \\ \hline
Human Example &1	&0.92&	1	&1&	1&	0.8&	1&	1&	1&	0.43&	1&	0.33\\ \hline

\end{tabular}
\caption{Unit Survival Rate (USR) Across Diverse Methods on the 12 HLSMAC Scenarios}
\label{tab:usr}
\end{table*}

\begin{figure*}[htbp]
\centering
\includegraphics[width=2\columnwidth]{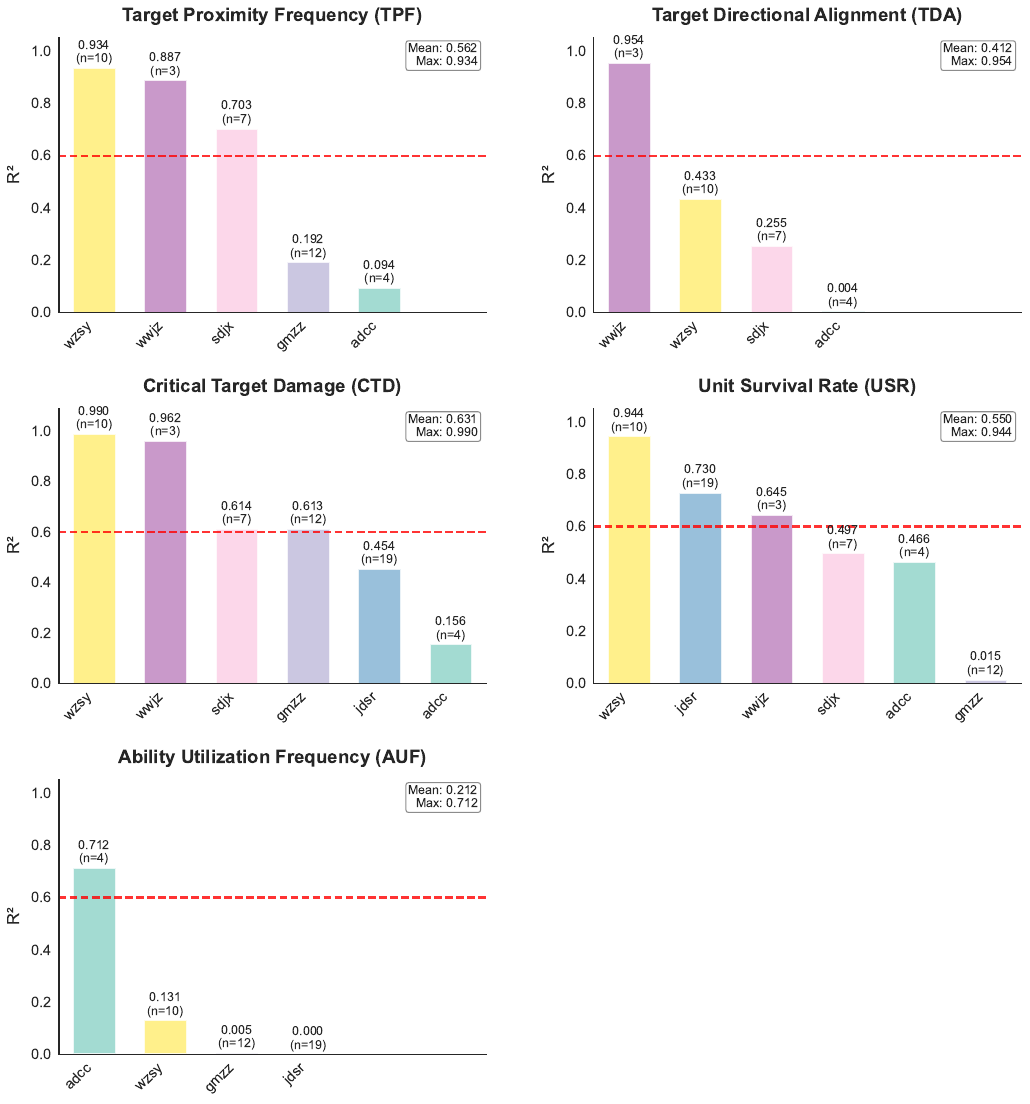} 
\caption{$R^2$ Analysis of Win Rate Correlation with Evaluation Metrics Across Scenarios. The correlation analysis reveals that Critical Target Damage (CTD) demonstrates the highest correlation with win rate, followed by Target Proximity Frequency (TPF) and Unit Survival Rate (USR). Target Directional Alignment (TDA) shows moderate correlation. This figure shows that Ability Utilization Frequency (AUF) has weaker correlation with win rate, likely because this frequency-based metric cannot fully reflect the appropriateness of ability usage. Nevertheless, AUF proves valuable for distinguishing human players from current methods, as it reveals that humans demonstrate purposeful utilization of critical abilities whereas current methods do not (see Table~\ref{tab:auf}). Note that metric-scenario combinations with win rates of zero were excluded during data preprocessing to maintain the validity of the correlation analysis.}
\label{fig:rsquaredanalysis}
\end{figure*}

\subsubsection{Results of LLM-PySC2 framework}
The following excerpts show how GPT-3.5 agents interpreted and responded to strategic scenarios. The LLM agents are unable to correctly comprehend the majority of scenarios, showing only limited tactical awareness in a few of them. For example, in the \textit{gmzz} scenario, the model correctly identifies that ``Our task is to use the strategy Shut The Door to Catch the Thief to defeat the Zerg army. We can utilize the Supply Depots to block the path of the Zerg units and trap them.'' Similarly, in the \textit{swct} scenario, it recognizes unit capabilities: ``Our team Sentry-1 is currently near the enemy Hatchery and has the ability to use ForceField to block enemy units.'' Additionally, agents exhibit ineffective action selection, failing to generate sequences that can successfully accomplish their stated goals.

\begin{itemize}
    \item \textbf{adcc}:	``...We need to use the stratagem `Openly repair the gallery roads, but sneak through the passage of Chencang' to deceive the enemy and launch a surprise attack.''
    \item \textbf{dhls}: ``Team Zergling-1 and Team Roach-1 are tasked with using the `Lure the Tiger Out of the Mountains' stratagem to defeat the enemy army...''
    \item \textbf{fkwz}:	``Both Team Zealot-1 and Team Zealot-2 have the task to stop the enemy from spawning an overwhelming army at the bottom left of the map. We need to focus on attacking the enemy buildings to disrupt their production.''
    \item \textbf{gmzz}:	``We are controlling a team called SupplyDepo-1, and we are facing a Zerg SpineCrawler nearby. Our task is to use the strategy Shut The Door to Catch the Thief to defeat the Zerg army. We can utilize the Supply Depots to block the path of the Zerg units and trap them.''
    \item \textbf{jctq}:	``... We need to create an illusion of doing one thing while actually doing another to successfully escape.''
    \item \textbf{jdsr}:	``Team Roach-1 and Team Infestor-1 are tasked with using the "Kill With a Borrowed Knife" stratagem to defeat the enemy army. The Roaches are tough ground units with short-range attacks, while the Infestors provide ground support and have powerful abilities like Fungal Growth and Neural Parasite.''
    \item \textbf{sdjx}:	``...To fulfill the tactical task of using Make a Feint to the East While Attacking in the West, we need to create a diversion in the east while launching a surprise attack in the west to destroy the enemy base....''
    \item \textbf{swct}:	``Our team Sentry-1 is currently near the enemy Hatchery and has the ability to use ForceField to block enemy units. We can utilize the stratagem "Pull Down the Ladder After the Ascent" to trap the enemy units near the Hatchery and prevent them from escaping.''
    \item \textbf{tlhz}:	``Our Drone is currently near an enemy Photon Cannon and Pylon. The Photon Cannon is low on health, and we have the opportunity to take it out before it gets repaired. It's crucial to eliminate this threat to secure our position and resources.''
    \item \textbf{wwjz}:	``Team Zealot-1 and Team Zealot-2 are tasked with defending the base against an enemy army. The enemy base is located at the bottom right of the map. We need to use the Besiege Wei to Rescue Zhao strategy to defeat the enemy forces.''
    \item \textbf{wzsy}:	``Our team, Sentry-1, is tasked with attacking the enemy base using the `Create something from nothing' stratagem. We have Sentries and nearby Stalkers for support. The previous action was to use ForceField at the enemy base. We need to continue the attack and create chaos in the enemy base.''
    \item \textbf{yqgz}: ``Our Zergling team is currently burrowed and ready for action. We have the opportunity to use the `In order to capture, one must let loose' tactic to defeat the enemy.''
\end{itemize}


% \end{document}

\end{document}